\documentclass[10pt,twocolumn,letterpaper]{article}

\usepackage{cvpr}              % To produce the CAMERA-READY version
\usepackage{graphicx}
\usepackage{amsmath}
\usepackage{amssymb}
\usepackage{booktabs}
\usepackage{multirow}

\usepackage[pagebackref,breaklinks,colorlinks]{hyperref}

\usepackage[capitalize]{cleveref}
\crefname{section}{Sec.}{Secs.}
\Crefname{section}{Section}{Sections}
\Crefname{table}{Table}{Tables}
\crefname{table}{Tab.}{Tabs.}

\def\cvprPaperID{*****} % *** Enter the CVPR Paper ID here
\def\confName{CVPR}
\def\confYear{2022}

\begin{document}

\title{ControlCom: Controllable Image Composition using Diffusion Model}

\author{
Bo Zhang$^{1}$
\quad Yuxuan Duan$^{1}$
\quad Jun Lan$^{2}$
\quad Yan Hong$^{2}$
\\
\quad Huijia Zhu$^{2}$
\quad Weiqiang Wang$^{2}$
\quad Li Niu\thanks{Corresponding author} $^{1}$ \\
\quad $^1$ Shanghai Jiao Tong University \\
{\tt\small\{bo-zhang, sjtudyx2016, ustcnewly\}@sjtu.edu.cn} \and
\quad $^2$ Ant Group \\
{\tt\small\{yelan.lj, ruoning.hy, huijia.zhj, weiqiang.wwq\}@antgroup.com}
}

\maketitle

%%%%%%%%% ABSTRACT
\begin{abstract}
   Image composition targets at synthesizing a realistic composite image from a pair of foreground and background images. Recently, generative composition methods are built on large pretrained diffusion models to generate composite images, considering their great potential in image generation. However, they suffer from lack of controllability on foreground attributes and poor preservation of foreground identity. To address these challenges, we propose a controllable image composition method that unifies four tasks in one diffusion model: image blending, image harmonization, view synthesis, and generative composition. Meanwhile, we design a self-supervised training framework coupled with a tailored pipeline of training data preparation. Moreover, we propose a local enhancement module to enhance the foreground details in the diffusion model, improving the foreground fidelity of composite images. The proposed method is evaluated on both public benchmark and real-world data, which demonstrates that our method can generate more faithful and controllable composite images than existing approaches. The code and model will be available at \url{https://github.com/bcmi/ControlCom-Image-Composition}.
\end{abstract}

%%%%%%%%% BODY TEXT
\section{Introduction}
\label{sec:introduction}

Image composition aims to synthesize a realistic composite image based on a foreground image with a desired object and a background image. To address the discrepancy  (\eg, illumination, pose) between foreground and background,  previous works decompose image composition into multiple tasks such as image blending, image harmonization, view synthesis, in which each task aims at solving one issue. Specifically, image blending~\cite{deepblending} copes with the unnatural boundary between foreground and background. Image harmonization~\cite{deepharmonization,dovenet} adjusts the foreground illumination to be compatible with background, while view synthesis~\cite{stgan} adjusts the foreground pose to be compatible with background. To get a realistic composite image with all issues solved, multiple models need to be applied sequentially, which is tedious and impractical. 

\begin{figure}[t]
    \begin{center}
    \includegraphics[width=1\linewidth]{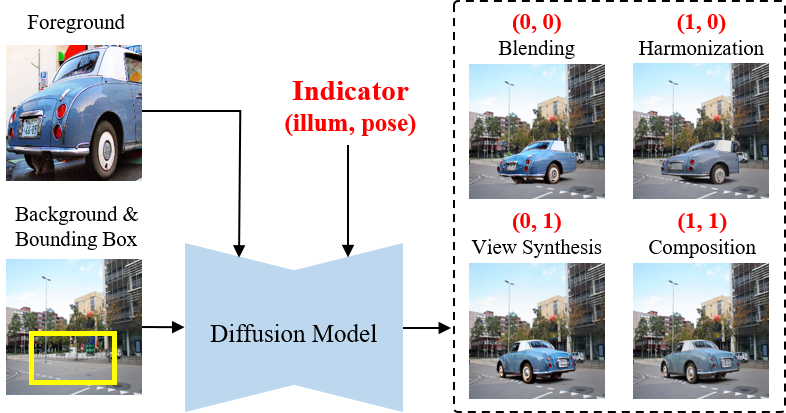}
    \end{center}
    \caption{Overview of our controllable image composition method. We unify four tasks in one diffusion model and enable control over the illumination and pose of the synthesized foreground objects with a 2-dim indicator vector.}
    \label{fig:task}
\end{figure}

Recently, generative image composition~\cite{objectstitch,paintbyexample} targets at solving all issues in one unified model, which can greatly simplify the composition pipeline. These methods are generally built on pretrained diffusion model \cite{stablediffusion}, due to its unprecedented power in synthesizing realistic images. Specifically, they take in a foreground image and a background image with a user-specified bounding box to produce a realistic composite image, in which a pretrained image encoder~\cite{clip} extracts the foreground embedding and the diffusion model incorporates this conditional foreground embedding into diffusion process.
However, these methods still suffer from lack of controllability and poor fidelity. Firstly, the diffusion model adjusts all attributes (\eg, illumination, pose) of foreground in an uncontrollable manner. Nevertheless, in some cases, some attributes (\eg, illumination, pose) of foreground are already compatible with background, and users may hope to preserve these attributes to avoid undesirable changes. Secondly, although the generated foreground belongs to the same semantic category as the input foreground, some appearance and texture details are dramatically altered, which does not satisfy the requirement of image composition. 

In this paper, we aim to address the above two issues: lack of controllability and poor fidelity. 
To address the first issue, we propose a \textbf{control}lable image \textbf{com}position method named ControlCom based on conditional diffusion model, which can selectively adjust partial foreground attributes (\ie, illumination, pose).
Specifically, we introduce a 2-dim indicator vector to indicate whether the illumination or pose of foreground should be changed. The indicator vector is injected into diffusion model as condition information. 
In this way, we unify four tasks in one model: image blending, image harmonization, view synthesis, generative composition (see Figure~\ref{fig:task}). 
When neither illumination nor pose is changed, our method performs image blending. When only illumination is changed, our method performs image harmonization. When only pose is changed, our method performs view synthesis. When both illumination and pose are changed, our method performs generative composition. 
We also design a self-supervised learning strategy to train four tasks simultaneously. 

To address the second issue, we extract both global embedding and local embeddings from foreground image. Different from previous methods~\cite{objectstitch,paintbyexample} which only fuse global embedding or fuse global/local embeddings simultaneously, we first fuse global embedding and then fuse local embeddings. By virtue of this two-stage fusion strategy, we can first generate the rough foreground object compatible with background, followed by filling in the appearance and texture details encapsulated in local embeddings. 
When fusing local embeddings, we also use aligned foreground embedding map constructed from local embeddings to refine the intermediate features in diffusion model, which contributes to more faithful depiction of the foreground object in the generated composite image.

We evaluate our proposed ControlCom on the public COCOEE dataset~\cite{paintbyexample}. We also build a real-world dataset named FOSCom upon existing dataset~\cite{discofos} for more comprehensive evaluation. Both qualitative and quantitative results demonstrate the superiority of our method in terms of both controllability and fidelity. Our major contribution can be summarized as follows: 
\begin{itemize}
\item  We propose a controllable image composition method that unifies four composition-related tasks with an indicator vector. We also design a self-supervised learning framework to train four tasks simultaneously. 
\item  We design a two-stage fusion strategy: first global fusion and then local fusion. In local fusion, we leverage the aligned foreground embedding map for feature modulation within diffusion model.
\item Extensive experiments on the public dataset and our dataset prove the effectiveness of our method. 
\end{itemize}

\section{Related Work}\label{sec:related_work}

\subsection{Image Composition}
The goal of image composition is combining the foreground object from one image and another background image to produce a composite image. However, the quality of composite images may suffer from the inconsistencies between foreground and background~\cite{compositionsurvey}, like unnatural boundary, inharmonious illumination, unsuitable pose, and so on. Since each issue is very challenging, several divergent tasks are proposed to solve each issue. 

Image blending~\cite{poissonblending,deepblending,zhang2021deep} paid attention to the elimination of unnatural boundary between foreground and background, which allows the foreground to be seamlessly blended into the background. Image harmonization~\cite{cdtnet,pctnet,dccf,dovenet,deepharmonization,xiaodong2019improving,sofiiuk2021foreground,regionaware,intriharm,guo2021image,bao2022deep,hang2022scs,jiang2021ssh,liang2021spatial,RenECCV2022,Harmonizer,WangCVPR2023,LEMaRT} focused on adjusting the illumination of foreground to match the background. However, despite the great success they have achieved in appearance adjustment, they cannot deal with the geometric inconsistency between foreground and background.
To cope with inconsistent camera viewpoint, several methods~\cite{stgan,sfgan,gccgan} were proposed to estimate warping parameters for the foreground for geometric correction. However, those methods typically predict affine or perspective transformation, which cannot handle complicated cases such as synthesizing foreground object with novel views or generalizing to non-rigid objects (\eg, person, animals). 

More recently, generative image composition \cite{paintbyexample,objectstitch,affordanceinsertion} targeted at solving all issues with one unified model and producing a composite image in an end-to-end manner. The representative works are PbE~\cite{paintbyexample} and ObjectStitch~\cite{objectstitch}. However, as discussed in Sec.~\ref{sec:introduction}, they suffer from low foreground fidelity and lack of control over the attributes of the synthesized foreground objects. 

\subsection{Subject-driven Image Generation and Editing}
Subject-driven image generation aims to generate images containing specific subjects situated in various scenes, while subject-driven image editing focuses on the replacement or incorporation of customized subjects into a given scene. Earlier works~\cite{interpretinggan,deepblending,hyperstyle} on subject-driven editing often leveraged the optimization and interpolation of GAN latent codes to generate particular subjects. With the advances of diffusion models, text-guided subject-driven image generation~\cite{textualinversion,dreambooth,customdiffusion} has received a growing attention, which primarily relies on enhanced textual prompts to synthesize the given subject. However, these approaches were computationally expensive due to the requirements of many iterations and multiple customized images for concept learning. For more efficient concept learning, recent studies~\cite{elite,encodertuning} attempted to develop specialized encoders to extract concept representation from customized images. Nevertheless, subject-driven image generation still struggles to control background scene and subject location. Besides, some efforts have been made in the direction of text-guided subject-driven image editing, \eg, subject swap~\cite{photoswap,customedit,blipdiffusion}. 

Different from the above studies, our work focuses on generative image composition, which can be viewed as a form of subject-driven image editing with precise control over subject location and background scene. Moreover, our method synthesizes composite images solely on foreground and background images without any textual input, and does not require finetuning at inference time.

\begin{figure*}[t]
    \begin{center}
    \includegraphics[width=1\linewidth]{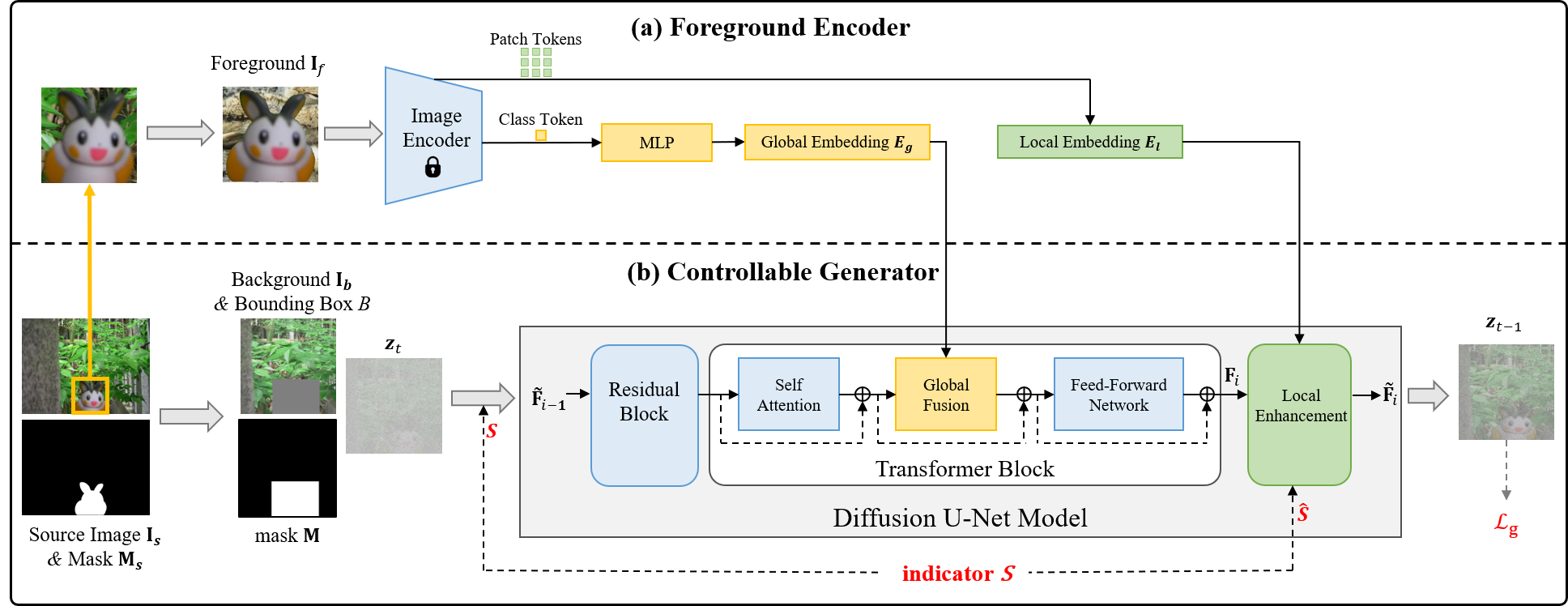}
    \end{center}
    \caption{Illustration of our ControlCom. Our model consists of two main components: a foreground encoder (a) that extracts hierarchical embeddings from foreground image, and a controllable generator (b) that synthesizes composite image with control over foreground illumination and pose using indicator $S$. See Figure~\ref{fig:local_enhance} for the details of local enhancement module. }
    \label{fig:overview_method}
\end{figure*}

\section{Preliminary}
\subsection{Problem Definition} 
Given an input tuple ($\mathbf{I}_b, \mathbf{I}_f, \mathbf{M}, B$) that consists of a background image $\mathbf{I}_b \in \mathbb{R}^{H_b \times W_b \times 3}$ with $H_b$ and $W_b$ being its height and width respectively, a foreground image $\mathbf{I}_f \in \mathbb{R}^{H_f \times W_f \times 3}$ containing the desired object (called foreground object) where $H_f$ and $W_f$ are the height and width of foreground image, a bounding box $B$ and its associated binary mask $\mathbf{M} \in \mathbb{R}^{H_b \times W_b \times 3}$ with the values within the bounding box being $1$, generative image composition aims to synthesize an image $\mathbf{I}_c$ that composites the foreground object into the background, so that the region within the bounding box depicts the object as similar to the foreground object and fits harmoniously, while the other regions remain as same as possible to the background $\mathbf{I}_b$. 

Furthermore, we introduce a 2-dim indicator vector $S$ to imply whether the illumination or pose of foreground should be changed during synthesizing composite images. In the indicator, the first dimension controls illumination and the second dimension controls pose, in which value $0$ (\emph{resp.}, $1$) means maintaining (\emph{resp.}, changing) the corresponding attribute of foreground. With this indicator, we can selectively adjust the foreground illumination and pose, which enables controllable image composition.

\subsection{Stable Diffusion}\label{sec:sd_model}
We build our controllable generator on a pretrained text-to-image diffusion model, Stable Diffusion (SD)~\cite{stablediffusion}. SD model consists of two components: an autoencoder $\{\mathcal{E}(\cdot), \mathcal{D}(\cdot)\}$ and a conditional diffusion model $\epsilon_{\theta}(\cdot)$. The autoencoder first maps an image $\mathbf{I} \in \mathbb{R}^{H \times W \times 3}$ to a lower dimensional latent space (\eg, $\mathbb{R}^{\frac{H}{8} \times \frac{W}{8} \times 4}$) by the encoder $\mathcal{E}(\cdot)$, and then maps the latent code $\mathcal{E}(\mathbf{I})$ back to the image by the decoder $\mathcal{D}(\cdot)$.
The conditional diffusion model $\epsilon_{\theta}(\cdot)$ is trained on the latent space to generate latent codes based on textual condition, whose architecture is implemented as a U-Net. The U-Net consists of a series of basic block with each block including a residual block and a transformer block. The transformer block consists of a self-attention module, a cross-attention module, and a feed-forward network. During forward computation, the features from the previous basic block first pass through the residual block and then are reorganized by the self-attention module. The resultant feature is fed into cross-attention module and interacts with the textual embedding from text prompt. This way, text prompts can be injected into the generated image.   

The conditional diffusion model generates image through a sequence of denoising steps, which is trained using the objective function~\cite{stablediffusion}:
\begin{equation}
    \label{eqn:loss_ldm}
    \mathcal{L}_{ldm} = \mathbb{E}_{\epsilon \sim \mathcal{N}(0,1), t} \| \epsilon - \epsilon_{\theta} \left( \mathbf{z}_t, \tau_{\theta}(y), t  \right) \|^2_2, 
\end{equation}
where $t$ is the time step that varies from 0 to $T$ and $\epsilon$ represents Gaussian noise. $\mathbf{z}_t$ represents the noisy version of the encoded image $\mathcal{E}(\mathbf{I})$ at time step $t$ and $\mathbf{z}_0 = \mathcal{E}(\mathbf{I})$. $y$ indicates text prompt and $\tau_{\theta}(\cdot)$ represents the pretrained CLIP encoder~\cite{clip}. With the noise estimation network, the reverse process can sample an image from random noise by gradually denoising it in $T$ time steps.

\section{Method}
\label{sec:method}
In this work, we propose a diffusion model based method for controllable image composition. 
As illustrated in Figure~\ref{fig:overview_method}, our method has two components: a foreground encoder $\phi_{\theta}(\cdot)$ and a controllable generator $\epsilon_{\theta}(\cdot)$. We use the foreground encoder to extract both global embedding encoding high-level semantics and local embeddings encoding fine-grained details from input foreground image $\mathbf{I}_f$ (Sec.~\ref{sec:fg_encoder}). With the foreground embedding as condition, we build controllable generator based on pretrained text-to-image diffusion model by replacing text prompt with foreground image. At inference time, the controllable generator takes the tuple ($\mathbf{I}_f, \mathbf{I}_b, \mathbf{M}, B, S$) as inputs and generates composite image $\mathbf{I}_c$ (Sec.~\ref{sec:controllable_generator}). To train the generator, we introduce several conditioning variables to the objective function in Eqn.~(\ref{eqn:loss_ldm}) and obtain the generator objective:
\begin{equation}
    \small
    \label{eqn:loss_ours}
    \mathcal{L}_{g} = \mathbb{E}_{\epsilon \sim \mathcal{N}(0,1), t} \| \epsilon - \epsilon_{\theta} \left( \mathbf{z}_t, \phi_{\theta}(\mathbf{I}_f), \mathcal{E}(\mathbf{I}_b),  \mathbf{M}, B, S, t \right) \|^2_2, 
\end{equation}
in which we include the encoded background image $\mathcal{E}(\mathbf{I}_b)$ generated by the encoder $\mathcal{E}(\cdot)$ of SD model and the mask $\mathbf{M}$ of bounding box $B$. Indicator $S$ is added to control the illumination or pose of foreground. Meanwhile, $\mathbf{z}_t$ changes to the noisy version of the encoded composite image $\mathcal{E}(\mathbf{I}_c)$ at time step $t$ and naturally $\mathbf{z}_0 = \mathcal{E}(\mathbf{I}_c)$. During inference, we first randomly sample a Gaussian noise as $\mathbf{z}_T$ and then iteratively denoise $\mathbf{z}_T$ to $\mathbf{z}_0$. Finally, we obtain composite image through the decoder as $\mathbf{I}^{'}_{c} = \mathcal{D}(\mathbf{z}_0)$. 

Moreover, we also design a self-supervised framework to train the proposed method on four tasks simultaneously, in which we collect training tuples ($\mathbf{I}_b, \mathbf{I}_f, \mathbf{M}, B, S, \mathbf{I}_c$) through the pipeline described in Sec.~\ref{sec:self-supervised framework}.   

\subsection{Foreground Encoder}\label{sec:fg_encoder}
An overview of the foreground encoder is shown in Figure~\ref{fig:overview_method} (a). We employ a pretrained ViT-L/14 image encoder from CLIP~\cite{clip} to extract feature from the resized foreground image $\mathbf{I}_f \in \mathbb{R}^{224 \times 224 \times 3}$. The intermediate layer of the CLIP encoder outputs 257 tokens with 1024 dimensions, including 1 class token that carries the high-level semantics and 256 patch tokens that contains local details. Following~\cite{paintbyexample}, we utilize the class token produced by the deepest encoder layer (\ie, layer 25) to yield global embedding $\mathbf{E}_g \in \mathbb{R}^{768}$ through a multilayer perceptron (MLP). However, only using class embedding cannot guarantee the preservation of foreground identity, due to lack of appearance and texture details of foreground. In order to enrich foreground details, we additionally fetch the patch tokens from a shallower layer (\ie, layer 12) as local embeddings $\mathbf{E}_l \in \mathbb{R}^{256 \times 1024}$. After that, we integrate both the global and local embeddings into intermediate features of diffusion model, yielding informative representations that enable more faithful foreground synthesis.

\subsection{Controllable Generator}\label{sec:controllable_generator}
We build our controllable generator on the publicly released v1-4 model of SD model. To fit our task, we append background $\mathbf{I}_b$ as well as binary mask $\mathbf{M}$ to the model input for easy reconstruction of background. Moreover, the indicator $S$ is used in two places: U-Net input and the proposed local enhancement module. To this end, we modify the conditional diffusion model by appending 7 additional channels to the first convolution layer in the U-Net. By unifying the resolution of input images to $512 \times 512$, the input of the U-Net has a shape of $64 \times 64 \times 11$. Among the $11$ channels, $4$ represents the noisy latent code $\mathbf{z}_t$, $4$ represents the encoded background image $\mathcal{E}(\mathbf{I}_b)$, $1$ represents the downsampled binary mask $\mathbf{M}'$, and $2$ represents indicator map $\mathbf{S}$ replicated from indicator vector $S$.  

To synthesize foreground object in composite image, we fuse both global and local embeddings of foreground into intermediate features of diffusion model. Instead of using global embedding and local embeddings simultaneously, we first fuse global embedding by the global fusion module and then fuse local embeddings by the local enhancement module (see Figure~\ref{fig:overview_method}). By means of the two-stage fusion strategy, we can first produce the rough foreground object consistent with background scene, followed by filling in the appearance and texture details attended from local embeddings, which promotes high-fidelity composite generation.

\noindent\textbf{Global Fusion.} Recall in Sec.~\ref{sec:sd_model}, the text-to-image diffusion model includes textual embedding into image generation leveraging the cross attention of U-Net. To cope with image composition, we replace the textual embedding with the global embedding of foreground, \ie, $\mathbf{E}_g$, which is injected to the intermediate representation of each transformer block in the U-Net through cross attention (see Figure~\ref{fig:overview_method}).

\begin{figure}[t]
    \begin{center}
    \includegraphics[width=1\linewidth]{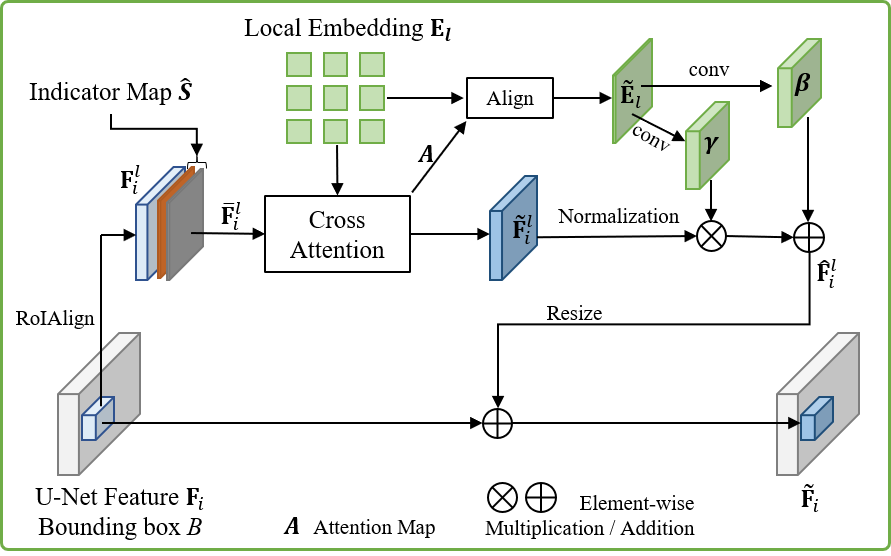}
    \end{center}
    \caption{Illustration of the local enhancement module.}
    \label{fig:local_enhance}
\end{figure}

\begin{figure*}[t]
    \begin{center}
    \includegraphics[width=1\linewidth]{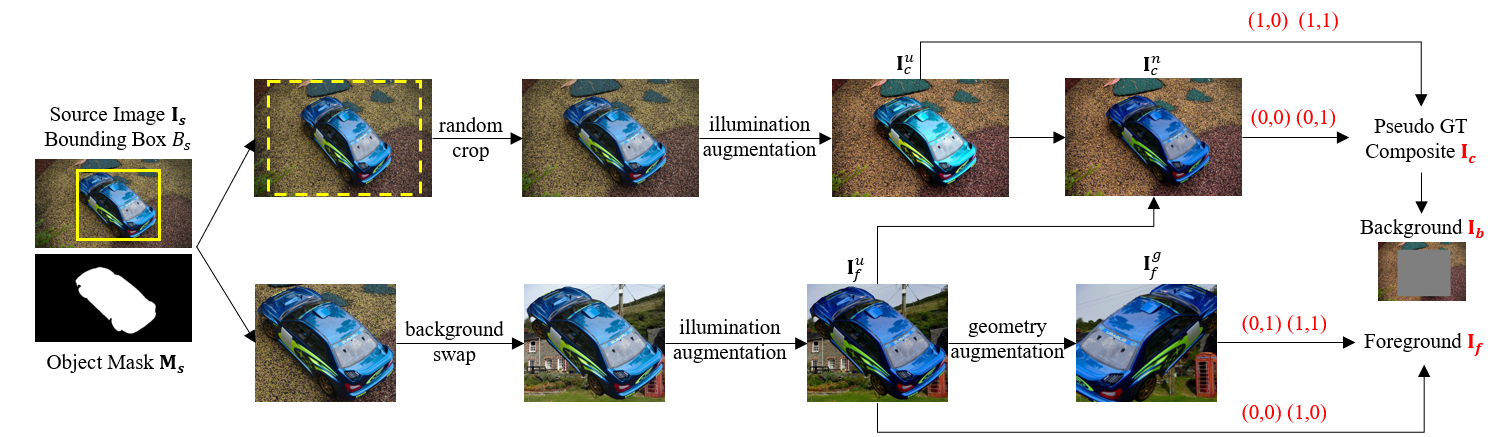}
    \end{center}
    \caption{Flowchart of synthetic data generation and augmentation.}
    \label{fig:data_pipeline}
\end{figure*}

\noindent\textbf{Local Enhancement.}
As illustrated in Figure~\ref{fig:local_enhance}, we denote the feature map produced by the $i$-th transformer block by $\mathbf{F}_i \in \mathbb{R}^{h_i \times w_i \times c_i}$ with $h_i, w_i, c_i$ being its height, width, and channel dimension. We first apply RoIAlign~\cite{roialign} with the bounding box $B$ to obtain local background feature $\mathbf{F}^l_i \in \mathbb{R}^{p \times p \times c_i}$ representing foreground region from $\mathbf{F}_i$. Here $p$ is the output size of RoIAlign. We then expand indicator $S$ to an indicator map $\hat{\mathbf{S}} \in \mathbb{R}^{p \times p \times 2}$ and concatenate it with the local background feature $\mathbf{F}^l_i$ to pass through a 3 $\times$ 3 convolutional layer for fusion. The resultant feature map is flattened to $\bar{\mathbf{F}}^l_{i} \in \mathbb{R}^{p^2 \times c_i}$, followed by cross attention with the local embeddings $\mathbf{E}_l$ of foreground. By fusing local embeddings $\mathbf{E}_l$, the local background feature $\bar{\mathbf{F}}^l_{i}$ can incorporate the fine-grained foreground information of $\mathbf{E}_l$ to generate foreground object more similar with the input one. After the cross attention, we obtain an attention map $\mathbf{A} \in \mathbb{R}^{p^2 \times 256}$ and the synthesized foreground feature map $\tilde{\mathbf{F}}^l_i \in \mathbb{R}^{p \times p \times c_i}$. 

To further boost the appearance and textual details of foreground, inspired by SPADE~\cite{SPADE}, we use an aligned foreground embedding map constructed from local embeddings $\mathbf{E}_l$ to modulate the synthesized foreground feature map $\tilde{\mathbf{F}}^l_i$. Specifically, the learned attention map $\mathbf{A}$ approximately captures the spatial correspondences between the input foreground and the synthesized foreground, so that we align the local embeddings $\mathbf{E}_l$ with $\tilde{\mathbf{F}}^l_i$ by multiplying $\mathbf{A}$ with $\mathbf{E}_l$ and reshaping the resultant embedding to 2D spatial structure, yielding aligned foreground embedding map $\tilde{\mathbf{E}}_l \in \mathbb{R}^{p \times p \times 1024}$. In the aligned foreground embedding map $\tilde{\mathbf{E}}_l$, each location has reasonable contextual information, which could help modulate $\tilde{\mathbf{F}}^l_i$. Thus, we perform convolution on $\tilde{\mathbf{E}}_l$ to get spatial-aware modulation weights for modulating the normalized $\tilde{\mathbf{F}}^l_i$ as follows,
\begin{equation}
    \hat{\mathbf{F}}^l_i = \mathrm{norm}(\tilde{\mathbf{F}}^l_i) \cdot \mathrm{conv}_\gamma(\tilde{\mathbf{E}}_l) + \mathrm{conv}_{\beta}(\tilde{\mathbf{E}}_l),
    \label{eqn:cond_norm}
\end{equation}
in which $\mathrm{conv}_\gamma$ and $\mathrm{conv}_\beta$ refer to 3 $\times$ 3 convolutional layers that convert aligned foreground embedding map into spatial-aware scale and shift modulation coefficients, respectively. The output $\hat{\mathbf{F}}^l_i$ has the same shape as the input feature $\tilde{\mathbf{F}}^l_i$. To integrate with the global background feature $\mathbf{F}_i$, we resize $\hat{\mathbf{F}}^l_i$ and add it to the bounding box region in $\mathbf{F}_i$, yielding enhanced feature $\tilde{\mathbf{F}}_i$ that is fed to the next residual block. The enhanced feature incorporates appearance and texture details of foreground, thereby improving the potential to generate high-fidelity composite images.

\subsection{Self-supervised Framework}
\label{sec:self-supervised framework}
Due to lack of dataset for training four tasks simultaneously, we propose a self-supervised framework together with a synthetic data preparation pipeline for learning these four tasks.

\noindent\textbf{Training Data Generation.}
We collect synthetic training data from a public large-scale dataset, \ie, Open Images~\cite{openimages}, which contains real-world images with object bounding boxes and partially with instance mask. We first filter the dataset and keep the objects with proper bounding box size (\emph{e.g.}, box area in 2\% $\sim$ 80\% of the area of whole image). Then we employ SAM~\cite{segmentanything} to predict instance mask for the objects without mask. After that, given a source image $\mathbf{I}_s$ and the bounding box $B_s$ in the image, we crop the bounding box enclosing the object as foreground image and mask out the bounding box region to create background image. To support controllable image composition, we design a data generation pipeline to produce training tuples (see Figure~\ref{fig:data_pipeline}).

\subsubsection{Training Data Augmentation.}
As shown in Figure~\ref{fig:data_pipeline}, we perform separate data augmentations to create composite image and foreground image. For composite image, we use both random crop and illumination augmentation to generate a variant of composite image, deemed as $\mathbf{I}^u_c$, in which we also update the bounding box $B_s$ to $B$. For the foreground image cropped from the same source image, we first apply background swap to replace the non-foreground region of foreground with other background, which prevents model from learning naive copy-and-paste. We then perform illumination augmentation and geometry augmentation on the foreground successively, producing $\mathbf{I}^u_f$ and $\mathbf{I}^g_f$, respectively. This procedure disturbs the illumination and pose of foreground, simulating practical scenarios where foreground has inconsistent illumination/pose with background. Finally, we replace the foreground object in $\mathbf{I}^u_c$ with that in $\mathbf{I}^u_f$ to get $\mathbf{I}^n_c$. 

\noindent\textbf{Training Sample Generation.}
With the transformed composites \{$\mathbf{I}^u_c$, $\mathbf{I}^n_c$\} and transformed foreground images \{$\mathbf{I}^u_f$, $\mathbf{I}^g_f$\}, we synthesize pseudo ground-truth (GT) composite image $\mathbf{I}_c$ and customize various training tuples for the four tasks. Each training tuple consists of background image, foreground image, bounding box and its corresponding binary mask, indicator, and pseudo ground-truth composite, \ie, ($\mathbf{I}_b$, $\mathbf{I}_f$, $B$, $S$, $\mathbf{M}$, $\mathbf{I}_c$). We assign each task with individual indicator $S$ and accordingly choose foreground image $\mathbf{I}_f$ as well as pseudo ground-truth composite $\mathbf{I}_c$. The principle is that the illumination or pose of foreground $\mathbf{I}_f$ should be consistent or inconsistent with $\mathbf{I}_c$ according to the indicator.

\begin{figure*}[t]
    \begin{center}
    \includegraphics[width=1\linewidth]{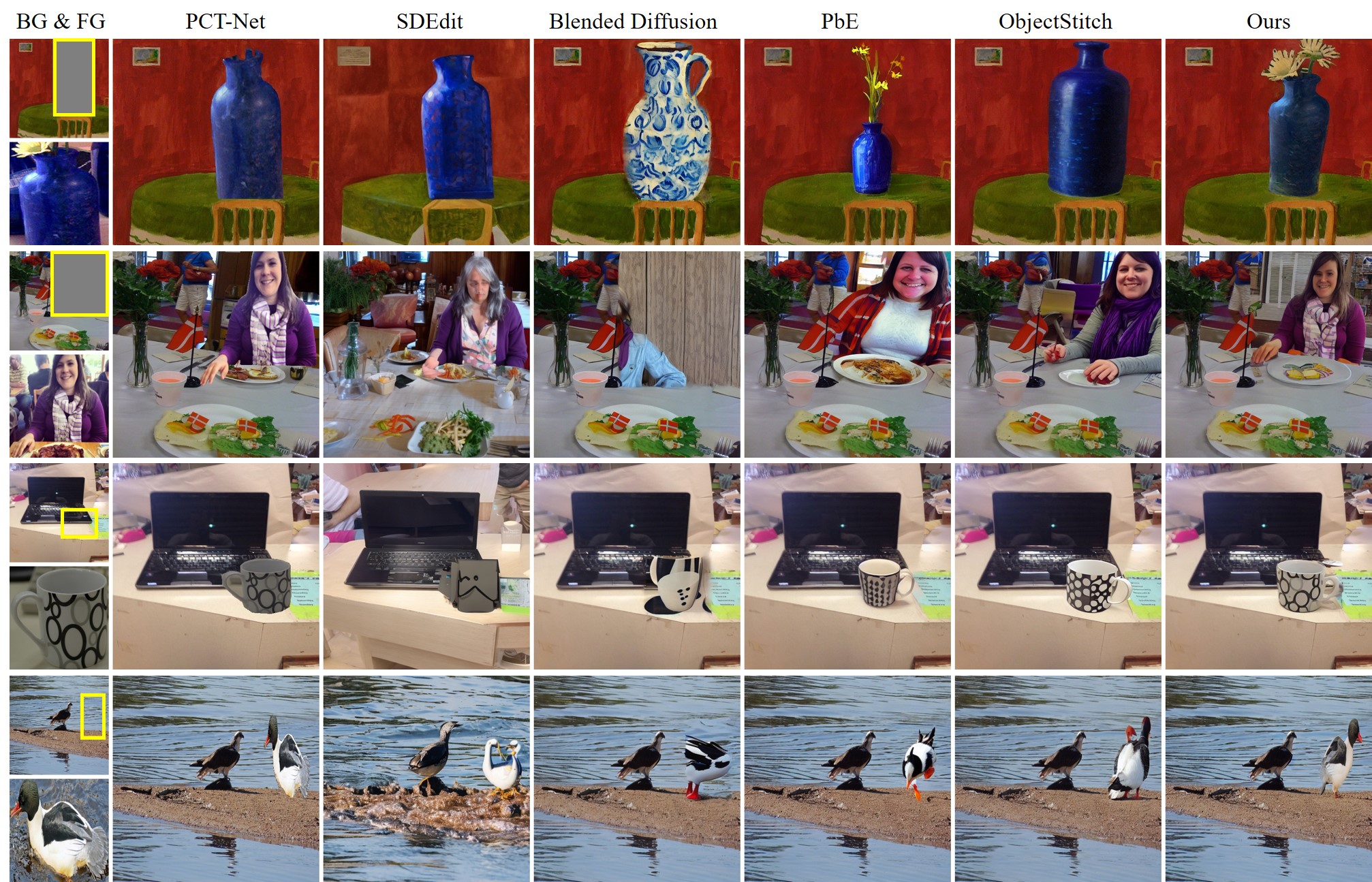}
    \end{center}
    \caption{Qualitative comparison on COCOEE dataset (top half) and our FOSCom dataset (bottom half). See Supp. for more visual results.}
    \label{fig:qualitative_comparison}
\end{figure*}

For image blending with indicator (0,0), we set $\mathbf{I}_f = \mathbf{I}^u_f, \mathbf{I}_c = \mathbf{I}^n_c$, in which the foreground object usually has inconsistent illumination with background scene. If $\mathbf{I}_c$ always has consistent illumination, then the model tends to alter the foreground illumination to fit the background during inference time. In contrast, by using inconsistent pseudo ground-truth composite images, we enforce model to maintain foreground illumination during image synthesis.
For image harmonization with indicator (1,0), we set $\mathbf{I}_f = \mathbf{I}^u_f, \mathbf{I}_c = \mathbf{I}^u_c$.
For view synthesis with indicator (0,1), we set $\mathbf{I}_f = \mathbf{I}^g_f, \mathbf{I}_c=\mathbf{I}^n_c$.
For generative composition with indicator (1,1), we set $\mathbf{I}_f = \mathbf{I}^g_f, \mathbf{I}_c = \mathbf{I}^u_c$. 
More details and explanations of training data preparation can be found in Supp. 

\section{Experiments}

\subsection{Implementation Details}
\label{sec:implement_details}
We implement our method using PyTorch~\cite{pytorch} and train our model on 16 NVIDIA A100 GPUs with random seed set as 23. The training set is built upon Open Images dataset~\cite{openimages} that contains 1.9 million images with 16 million objects covering 600 categories. We use the Adam optimizer~\cite{adam} with a fixed learning rate of $1e^{-5}$ and batch size of 256 to train our model for 40 epochs. 

In our ControlCom, the MLP of foreground encoder consists of five fully-connected layers. In the local enhancement module, we set the output size of RoIAlign~\cite{roialign} as 16, \ie, $p=16$. To avoid information loss of low-resolution feature maps, we append local enhancement module to the transformer blocks that output feature map with spatial resolution of 64 $\times$ 64 or 32 $\times$ 32. The results of our method are generated with indicator (1,1). Moreover, we adopt classifier-free sampling strategy~\cite{classifierfreeguidance} to improve image quality following~\cite{paintbyexample}. Specifically, we replace 20\% global embedding $\mathbf{E}_g$ with a learnable embedding during training and set the classifier-free guidance scale to 5 for sampling. During inference, we use DDIM sampler~\cite{ddim} with 50 steps for our method and other baselines. 

\subsection{Dataset and Evaluation Metrics}
\label{sec:dataset_metrics}
\noindent\textbf{COCOEE Dataset.} COCO Exemplar-based image Editing benchmark (COCOEE)~\cite{paintbyexample} has 3500 pairs of background and foreground, which are respectively collected from the validation set and training set of COCO dataset~\cite{coco}. Each background image has a bounding box and the masked region shares similar semantics with foreground image, which ensures the plausibility of the composite results.

\noindent\textbf{FOSCom Dataset.} 
In COCOEE~\cite{paintbyexample}, the foreground region of background image always contains certain object, which may degrade the realism of composite images. In contrast, real-world image composition typically inserts one object to an open area. 
For real-world evaluation, we construct a dataset named FOSCom based on existing Foreground Object Search (FOS) dataset~\cite{discofos}, which contains 640 background images from Internet. Each background image has an manually annotated bounding box, which is suitable to place one object from a specified category. To adapt this dataset to our task, we collect one foreground image for each background image from the training set of COCO dataset~\cite{coco}. The resultant dataset has 640 pairs of backgrounds and foregrounds, which is used in our user study and qualitative comparison.   

\noindent\textbf{Evaluation Metrics.} Following~\cite{paintbyexample}, we adopt Fréchet Inception Distance (FID)~\cite{fid} and Quality Score (QS)~\cite{qualityscore} to evaluate the authenticity of generated composite images, in which FID and QS are calculated between synthesized images and all images from the test set of COCOEE dataset~\cite{paintbyexample}.
For foreground, we employ CLIP score~\cite{clip} ($\mathrm{CLIP_{fg}}$) to indicate generative fidelity, \ie, the similarity between the given and synthesized foreground object. Specifically, we crop out foreground patch from generated composite and mask the non-object region to compute CLIP score with the masked input foreground, in which we estimate the mask of foreground object using SAM~\cite{segmentanything}.
For background, we leverage SSIM~\cite{ssim} and LPIPS~\cite{lpips} to assess reconstruction error, in which we fill the bounding boxes of input backgrounds and generated composites with black.

\subsection{Comparisons}
\noindent\textbf{Baselines.} To investigate the effectiveness of our method, we compare with various baselines (see Table~\ref{tab:baseline}). Among these baselines, Inpaint\&Paste is implemented by first inpainting the foreground object area of background using Stable Diffusion~\cite{stablediffusion} and then pasting foreground object onto the filled background. We consider the image produced by Inpaint\&Paste as naive composite results. PCT-Net~\cite{pctnet} is a harmonization baseline used to harmonize the naive composite results. For text-guided image generation/editing methods, \ie, SDEdit~\cite{sdedit} and Blended Diffusion~\cite{blendedlatentdiffusion} (BlendedDiff), we utilize BLIP~\cite{blip} to produce captions for composite and foreground images to guide image synthesis. ObjectStitch~\cite{objectstitch} and PbE~\cite{paintbyexample} are generative composition methods, which share the same inputs and pretrained diffusion model to ours.

\noindent\textbf{Quantitative Comparison.} In Table~\ref{tab:baseline}, we evaluate different approaches on COCOEE dataset and report the results of the evaluation metrics introduced in Sec.~\ref{sec:dataset_metrics}. 
Among these baselines, PCT-Net achieves the best results in respect of $\mathrm{CLIP_{fg}}$, as it harmonizes the composite image produced by copy-and-paste. Blended Diffusion~\cite{blendedlatentdiffusion} performs strongly in background preservation, owing to directly blending the noisy version of the encoded background image with the local foreground latent. However, both of them are prone to produce unnatural or implausible image with poor quality. Regarding overall quality of generated composites, PbE~\cite{paintbyexample} is the most competitive baseline to our model. Compared with PbE~\cite{paintbyexample} and ObjectStitch~\cite{objectstitch}, all of our four versions achieve better results on foreground fidelity, and our Composition version achieves comparable performance on background preservation and generative quality.

\begin{table}
\centering
\resizebox{\linewidth}{!}{
\begin{tabular}{l|c|cc|cc}
\hline
\multirow{2}*{Method} & \multicolumn{1}{c|}{Foreground} &\multicolumn{2}{c|}{Background} &\multicolumn{2}{c}{Overall}\\
&$\mathrm{CLIP_{fg}}$↑   &LSSIM↑ &LPIPS↓   &FID↓   &QS↑ \\ 
\hline \hline
Inpaint\&Paste &8.0 &-- &-- &3.64 &72.07 \\ 
PCT-Net &\textbf{\underline{99.15}} &-- &-- &3.53 &72.81\\ 
SDEdit &85.02 &0.630 &0.344 &6.42 &75.20 \\  
BlendedDiff &76.62 &\textbf{\underline{0.833}} &\textbf{\underline{0.112}} &3.95 &71.53\\
ObjectStitch &85.97 &0.825 &0.116 &3.35 &76.86 \\
PbE &84.84 &0.823 &0.116 &\textbf{\underline{3.18}} &\textbf{77.80} \\
\hline
Ours (Blend) &\textbf{90.63} &0.826 &0.114 &3.25 &77.38 \\
Ours (Harm) &90.59 &0.826 &0.114 &3.22 &77.77 \\
Ours (View) &88.38 &0.826 &0.114 &3.23 &77.41 \\
Ours (Comp) &88.31 &\textbf{0.826} &\textbf{0.114} &\textbf{3.19} &\textbf{\underline{77.84}} \\
\hline
\end{tabular}
}
\caption{Quantitative comparison on COCOEE dataset. Blend: image blending. Harm: image harmonization. View: view synthesis. Comp: generative composition. ``--'': meaningless entries. Boldface denotes the best two results and underline highlights the best one.}
\label{tab:baseline}
\end{table}

Among the four versions, our Blending version and Harmonization version achieve higher $\mathrm{CLIP_{fg}}$, as both tasks tend to directly copy and paste original/harmonized foreground object on the background. Concurrently, the quality of generated images will be constrained due to incomplete boundary or incompatible pose of foreground. For our ViewSynthesis version, the quality of our generated images may be limited by inharmonious illumination and its fidelity may be harmed by deviation from the pose of input foreground. In contrast, by virtue of adjusting both illumination and pose of foreground, our Composition version outputs more plausible composite images, significantly increasing the overall quality of generated composite images.

\noindent\textbf{Qualitative Comparison.} In Figure~\ref{fig:qualitative_comparison}, we provide visual comparison results of different methods on both COCOEE and FOSCom datasets. For each example, we display background, foreground, and the composite images produced by different methods. It can be seen that PCT-Net obtains almost the same as the foreground image, which is very incongruous with the background. The text-guided diffusion models, including SDEdit~\cite{sdedit} and Blended Diffusion~\cite{blendedlatentdiffusion}, can output much plausible results, yet diverge from input foreground due to the limited representation of text information. The results of generative composition models (PbE~\cite{paintbyexample} and ObjectStitch~\cite{objectstitch}) clearly improve both overall quality and foreground similarity, but still fail to retain the details of foreground. In contrast, our model can generally yield realistic composite images while preserving foreground appearance and texture details. More qualitative results are shown in Supp.

\begin{figure}[t]
    \begin{center}
    \includegraphics[width=1\linewidth]{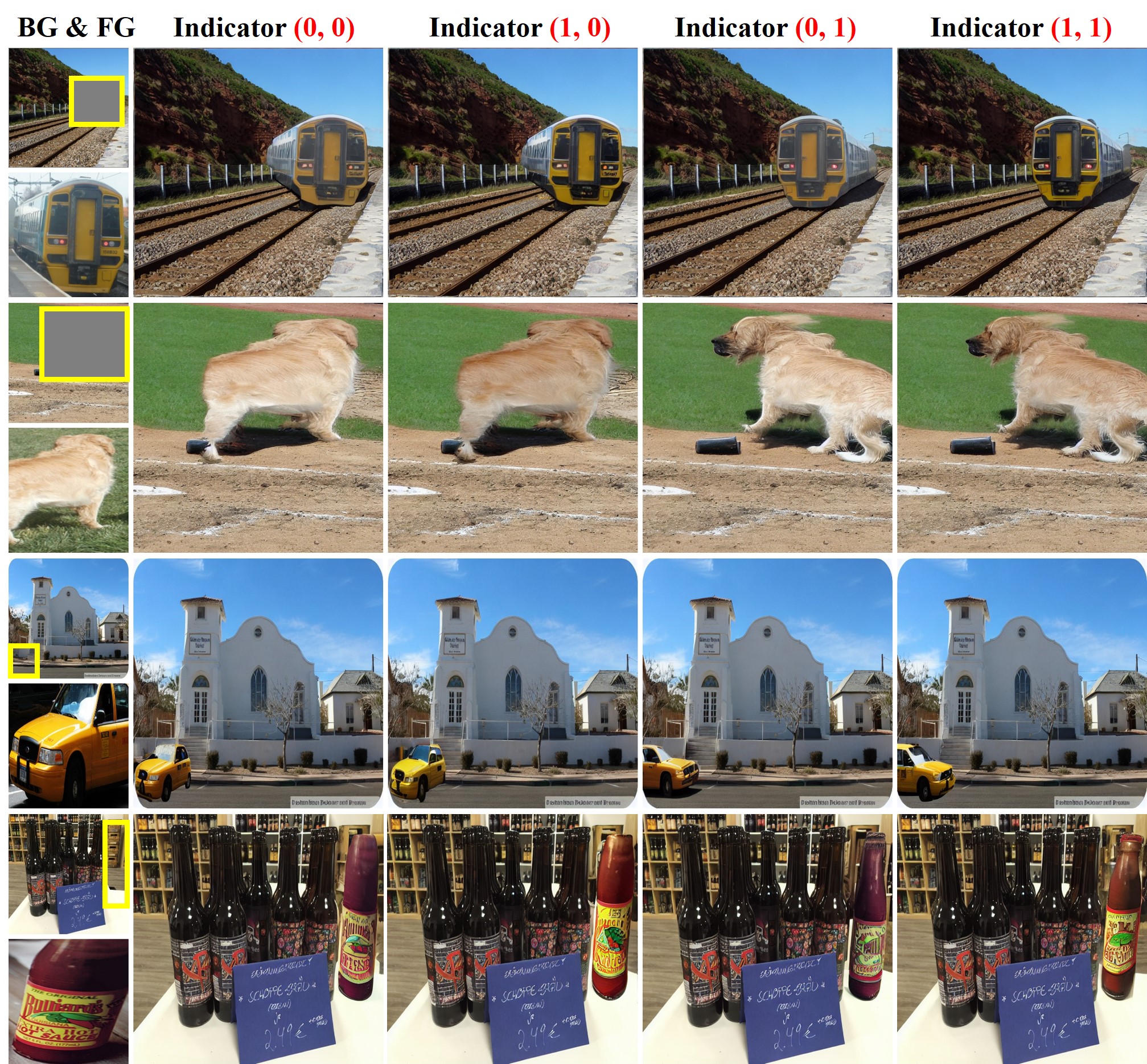}
    \end{center}
    \caption{Controllable image composition on COCOEE dataset (top half) and our FOSCom dataset (bottom half).}
    \label{fig:controllable_results}
\end{figure}

\subsection{Analyses on Individual Tasks}
To demonstrate the controllability of our method, we provide the visual results of our four versions in Figure~\ref{fig:controllable_results}, in which four composite images are sampled from the same initial noise using different indicators. In the second column, foreground is seamlessly pasted on the background, while retaining both illumination and pose of foreground. In the third column, our model can adaptively adjust foreground illumination, generating composite image with harmonious appearance. In the last columns, the composite images are much realistic, owing to synthesizing novel viewpoint of foreground and automatically completing the incomplete boundary. Thus we can confirm the controllability of our method over foreground illumination and pose. We also conduct user studies on individual tasks, \ie, image blending, image harmonization, and generative composition, which are left in Supp.  
\subsection{Additional Experiments in Supplementary}
\label{sec:additional_experiments}
Due to space limitation, we present some experiments in Supp., including ablation study of our method, user studies on individual tasks, more visualization results, and limitation discussion of our method.

\section{Conclusion}
In this work, we have proposed a controllable image composition method that unifies four tasks in one diffusion model. Equiped with a tailored pipeline of training data preparation, we trained our method in a self-supervised framework. We have also proposed a two-stage fusion strategy for conditioning diffusion model on foreground image. Extensive experiments on the pubic dataset and our dataset have validated the proposed method on generative image composition.

%%%%%%%%% REFERENCES
{\small
\bibliographystyle{ieee_fullname}
\bibliography{egbib}

\begin{thebibliography}{10}\itemsep=-1pt

\bibitem{hyperstyle}
Yuval Alaluf, Omer Tov, Ron Mokady, Rinon Gal, and Amit~H. Bermano.
\newblock {HyperStyle}: Stylegan inversion with hypernetworks for real image
  editing.
\newblock {\em CVPR}, 2021.

\bibitem{blendedlatentdiffusion}
Omri Avrahami, Ohad Fried, and Dani Lischinski.
\newblock Blended latent diffusion.
\newblock {\em ArXiv}, abs/2206.02779, 2022.

\bibitem{bao2022deep}
Zhongyun Bao, Chengjiang Long, Gang Fu, Daquan Liu, Yuanzhen Li, Jiaming Wu,
  and Chunxia Xiao.
\newblock Deep image-based illumination harmonization.
\newblock In {\em CVPR}, 2022.

\bibitem{gccgan}
Bor{-}Chun Chen and Andrew Kae.
\newblock Toward realistic image compositing with adversarial learning.
\newblock In {\em CVPR}, 2019.

\bibitem{customedit}
Jooyoung Choi, Yunjey Choi, Yunji Kim, Junho Kim, and Sung-Hoon Yoon.
\newblock Custom-edit: Text-guided image editing with customized diffusion
  models.
\newblock {\em ArXiv}, abs/2305.15779, 2023.

\bibitem{cdtnet}
Wenyan Cong, Xinhao Tao, Li Niu, Jing Liang, Xuesong Gao, Qihao Sun, and Liqing
  Zhang.
\newblock High-resolution image harmonization via collaborative dual
  transformations.
\newblock In {\em CVPR}, 2022.

\bibitem{dovenet}
Wenyan Cong, Jianfu Zhang, Li Niu, Liu Liu, Zhixin Ling, Weiyuan Li, and Liqing
  Zhang.
\newblock {DoveNet}: Deep image harmonization via domain verification.
\newblock In {\em CVPR}, 2020.

\bibitem{xiaodong2019improving}
Xiaodong Cun and Chi{-}Man Pun.
\newblock Improving the harmony of the composite image by spatial-separated
  attention module.
\newblock {\em {IEEE} Transactions on Image Processing}, 29:4759--4771, 2020.

\bibitem{textualinversion}
Rinon Gal, Yuval Alaluf, Yuval Atzmon, Or Patashnik, Amit~H. Bermano, Gal
  Chechik, and Daniel Cohen-Or.
\newblock An image is worth one word: Personalizing text-to-image generation
  using textual inversion.
\newblock {\em ArXiv}, abs/2208.01618, 2022.

\bibitem{encodertuning}
Rinon Gal, Moab Arar, Yuval Atzmon, Amit~H. Bermano, Gal Chechik, and Daniel
  Cohen-Or.
\newblock Encoder-based domain tuning for fast personalization of text-to-image
  models.
\newblock {\em ArXiv}, abs/2302.12228, 2023.

\bibitem{photoswap}
Jing Gu, Yilin Wang, Nanxuan Zhao, Tsu-Jui Fu, Wei Xiong, Qing Liu, Zhifei
  Zhang, He Zhang, Jianming Zhang, Hyun-Sun Jung, and Xin Wang.
\newblock Photoswap: Personalized subject swapping in images.
\newblock {\em ArXiv}, abs/2305.18286, 2023.

\bibitem{qualityscore}
Shuyang Gu, Jianmin Bao, Dong Chen, and Fang Wen.
\newblock Giqa: Generated image quality assessment.
\newblock In {\em ECCV}, 2020.

\bibitem{pctnet}
Julian Jorge~Andrade Guerreiro, Mitsuru Nakazawa, and Bj\"orn Stenger.
\newblock Pct-net: Full resolution image harmonization using pixel-wise color
  transformations.
\newblock In {\em CVPR}, 2023.

\bibitem{guo2021image}
Zonghui Guo, Dongsheng Guo, Haiyong Zheng, Zhaorui Gu, Bing Zheng, and Junyu
  Dong.
\newblock Image harmonization with transformer.
\newblock In {\em ICCV}, 2021.

\bibitem{intriharm}
Zonghui Guo, Haiyong Zheng, Yufeng Jiang, Zhaorui Gu, and Bing Zheng.
\newblock Intrinsic image harmonization.
\newblock In {\em CVPR}, 2021.

\bibitem{hang2022scs}
Yucheng Hang, Bin Xia, Wenming Yang, and Qingmin Liao.
\newblock {SCS-C}o: Self-consistent style contrastive learning for image
  harmonization.
\newblock In {\em CVPR}, 2022.

\bibitem{roialign}
Kaiming He, Georgia Gkioxari, Piotr Doll{\'a}r, and Ross~B. Girshick.
\newblock {Mask R-CNN}.
\newblock {\em PAMI}, 42:386--397, 2020.

\bibitem{fid}
Martin Heusel, Hubert Ramsauer, Thomas Unterthiner, Bernhard Nessler, and Sepp
  Hochreiter.
\newblock Gans trained by a two time-scale update rule converge to a local nash
  equilibrium.
\newblock In {\em NIPS}, 2017.

\bibitem{classifierfreeguidance}
Jonathan Ho.
\newblock Classifier-free diffusion guidance.
\newblock {\em ArXiv}, abs/2207.12598, 2022.

\bibitem{jiang2021ssh}
Yifan Jiang, He Zhang, Jianming Zhang, Yilin Wang, Zhe Lin, Kalyan Sunkavalli,
  Simon Chen, Sohrab Amirghodsi, Sarah Kong, and Zhangyang Wang.
\newblock Ssh: A self-supervised framework for image harmonization.
\newblock In {\em ICCV}, 2021.

\bibitem{Harmonizer}
Zhanghan Ke, Chunyi Sun, Lei Zhu, Ke Xu, and Rynson~W.H. Lau.
\newblock Harmonizer: Learning to perform white-box image and video
  harmonization.
\newblock In {\em ECCV}, 2022.

\bibitem{adam}
Diederik~P Kingma and Jimmy Ba.
\newblock Adam: A method for stochastic optimization.
\newblock In {\em ICLR}, 2015.

\bibitem{segmentanything}
Alexander Kirillov, Eric Mintun, Nikhila Ravi, Hanzi Mao, Chloe Rolland, Laura
  Gustafson, Tete Xiao, Spencer Whitehead, Alexander~C. Berg, Wan-Yen Lo, Piotr
  Doll{\'a}r, and Ross Girshick.
\newblock Segment anything.
\newblock {\em arXiv:2304.02643}, 2023.

\bibitem{affordanceinsertion}
Sumith Kulal, Tim Brooks, Alex Aiken, Jiajun Wu, Jimei Yang, Jingwan Lu,
  Alexei~A. Efros, and Krishna~Kumar Singh.
\newblock Putting people in their place: Affordance-aware human insertion into
  scenes.
\newblock In {\em CVPR}, 2023.

\bibitem{customdiffusion}
Nupur Kumari, Bin Zhang, Richard Zhang, Eli Shechtman, and Jun-Yan Zhu.
\newblock Multi-concept customization of text-to-image diffusion.
\newblock In {\em CVPR}, 2023.

\bibitem{openimages}
Alina Kuznetsova, Hassan Rom, Neil~Gordon Alldrin, Jasper R.~R. Uijlings, Ivan
  Krasin, Jordi Pont-Tuset, Shahab Kamali, Stefan Popov, Matteo Malloci,
  Alexander Kolesnikov, Tom Duerig, and Vittorio Ferrari.
\newblock The open images dataset v4.
\newblock {\em IJCV}, 128:1956--1981, 2018.

\bibitem{blipdiffusion}
Dongxu Li, Junnan Li, and Steven Hoi.
\newblock Blip-diffusion: Pre-trained subject representation for controllable
  text-to-image generation and editing.
\newblock {\em ArXiv}, abs/2305.14720, 2023.

\bibitem{blip}
Junnan Li, Dongxu Li, Caiming Xiong, and Steven Hoi.
\newblock Blip: Bootstrapping language-image pre-training for unified
  vision-language understanding and generation.
\newblock In {\em ICML}, 2022.

\bibitem{liang2021spatial}
Jingtang Liang, Xiaodong Cun, and Chi-Man Pun.
\newblock Spatial-separated curve rendering network for efficient and
  high-resolution image harmonization.
\newblock {\em ECCV}, 2022.

\bibitem{stgan}
Chen{-}Hsuan Lin, Ersin Yumer, Oliver Wang, Eli Shechtman, and Simon Lucey.
\newblock {ST-GAN:} spatial transformer generative adversarial networks for
  image compositing.
\newblock In {\em CVPR}, 2018.

\bibitem{coco}
Tsung-Yi Lin, Michael Maire, Serge Belongie, James Hays, Pietro Perona, Deva
  Ramanan, Piotr Doll{\'a}r, and C~Lawrence Zitnick.
\newblock Microsoft coco: Common objects in context.
\newblock In {\em ECCV}, 2014.

\bibitem{regionaware}
Jun Ling, Han Xue, Li Song, Rong Xie, and Xiao Gu.
\newblock Region-aware adaptive instance normalization for image harmonization.
\newblock In {\em CVPR}, 2021.

\bibitem{LEMaRT}
Sheng Liu, Cong~Phuoc Huynh, Cong Chen, Maxim Arap, and Raffay Hamid.
\newblock Lemart: Label-efficient masked region transform for image
  harmonization.
\newblock In {\em CVPR}, 2023.

\bibitem{sdedit}
Chenlin Meng, Yutong He, Yang Song, Jiaming Song, Jiajun Wu, Jun-Yan Zhu, and
  Stefano Ermon.
\newblock Sdedit: Guided image synthesis and editing with stochastic
  differential equations.
\newblock In {\em ICLR}, 2021.

\bibitem{compositionsurvey}
Li Niu, Wenyan Cong, Liu Liu, Yan Hong, Bo Zhang, Jing Liang, and Liqing Zhang.
\newblock Making images real again: A comprehensive survey on deep image
  composition.
\newblock {\em ArXiv}, abs/2106.14490, 2021.

\bibitem{SPADE}
Taesung Park, Ming-Yu Liu, Ting-Chun Wang, and Jun-Yan Zhu.
\newblock Semantic image synthesis with spatially-adaptive normalization.
\newblock In {\em CVPR}, 2019.

\bibitem{pytorch}
Adam Paszke, Sam Gross, Francisco Massa, Adam Lerer, James Bradbury, Gregory
  Chanan, Trevor Killeen, Zeming Lin, Natalia Gimelshein, Luca Antiga, et~al.
\newblock {PyTorch}: An imperative style, high-performance deep learning
  library.
\newblock In {\em NIPS}, 2019.

\bibitem{poissonblending}
Patrick P{\'e}rez, Michel Gangnet, and Andrew Blake.
\newblock Poisson image editing.
\newblock {\em SIGGRAPH}, 2003.

\bibitem{clip}
Alec Radford, Jong~Wook Kim, Chris Hallacy, Aditya Ramesh, Gabriel Goh,
  Sandhini Agarwal, Girish Sastry, Amanda Askell, Pamela Mishkin, Jack Clark,
  Gretchen Krueger, and Ilya Sutskever.
\newblock Learning transferable visual models from natural language
  supervision.
\newblock In {\em ICML}, 2021.

\bibitem{RenECCV2022}
Xuqian Ren and Yifan Liu.
\newblock Semantic-guided multi-mask image harmonization.
\newblock In {\em ECCV}, 2022.

\bibitem{stablediffusion}
Robin Rombach, A. Blattmann, Dominik Lorenz, Patrick Esser, and Bj{\"o}rn
  Ommer.
\newblock High-resolution image synthesis with latent diffusion models.
\newblock In {\em CVPR}, 2022.

\bibitem{dreambooth}
Nataniel Ruiz, Yuanzhen Li, Varun Jampani, Yael Pritch, Michael Rubinstein, and
  Kfir Aberman.
\newblock Dreambooth: Fine tuning text-to-image diffusion models for
  subject-driven generation.
\newblock In {\em CVPR}, 2023.

\bibitem{interpretinggan}
Yujun Shen, Jinjin Gu, Xiaoou Tang, and Bolei Zhou.
\newblock Interpreting the latent space of gans for semantic face editing.
\newblock {\em CVPR}, 2019.

\bibitem{sofiiuk2021foreground}
Konstantin Sofiiuk, Polina Popenova, and Anton Konushin.
\newblock Foreground-aware semantic representations for image harmonization.
\newblock In {\em WACV}, 2021.

\bibitem{ddim}
Jiaming Song, Chenlin Meng, and Stefano Ermon.
\newblock Denoising diffusion implicit models.
\newblock In {\em ICLR}, 2020.

\bibitem{objectstitch}
Yi-Zhe Song, Zhifei Zhang, Zhe~L. Lin, Scott~D. Cohen, Brian~L. Price, Jianming
  Zhang, Soo~Ye Kim, and Daniel~G. Aliaga.
\newblock Objectstitch: Generative object compositing.
\newblock In {\em CVPR}, 2023.

\bibitem{deepharmonization}
Yi{-}Hsuan Tsai, Xiaohui Shen, Zhe Lin, Kalyan Sunkavalli, Xin Lu, and
  Ming{-}Hsuan Yang.
\newblock Deep image harmonization.
\newblock In {\em CVPR}, 2017.

\bibitem{WangCVPR2023}
Ke Wang, Michaël Gharbi, He Zhang, Zhihao Xia, and Eli Shechtman.
\newblock Semi-supervised parametric real-world image harmonization.
\newblock In {\em CVPR}, 2023.

\bibitem{ssim}
Zhou Wang, Alan~C Bovik, Hamid~R Sheikh, and Eero~P Simoncelli.
\newblock Image quality assessment: from error visibility to structural
  similarity.
\newblock {\em TIP}, 13(4):600--612, 2004.

\bibitem{elite}
Yuxiang Wei, Yabo Zhang, Zhilong Ji, Jinfeng Bai, Lei Zhang, and Wangmeng Zuo.
\newblock {ELITE}: Encoding visual concepts into textual embeddings for
  customized text-to-image generation.
\newblock {\em ArXiv}, abs/2302.13848, 2023.

\bibitem{dccf}
Ben Xue, Shenghui Ran, Quan Chen, Rongfei Jia, Binqiang Zhao, and Binqiang
  Zhao.
\newblock Dccf: Deep comprehensible color filter learning framework for
  high-resolution image harmonization.
\newblock In {\em ECCV}, 2022.

\bibitem{paintbyexample}
Binxin Yang, Shuyang Gu, Bo Zhang, Ting Zhang, Xuejin Chen, Xiaoyan Sun, Dong
  Chen, and Fang Wen.
\newblock Paint by example: Exemplar-based image editing with diffusion models.
\newblock In {\em CVPR}, 2023.

\bibitem{sfgan}
Fangneng Zhan, Hongyuan Zhu, and Shijian Lu.
\newblock Spatial fusion {GAN} for image synthesis.
\newblock In {\em CVPR}, 2019.

\bibitem{discofos}
Bo Zhang, Jiacheng Sui, and Li Niu.
\newblock Foreground object search by distilling composite image feature.
\newblock {\em ArXiv}, abs/2308.04990, 2023.

\bibitem{zhang2021deep}
He Zhang, Jianming Zhang, Federico Perazzi, Zhe Lin, and Vishal~M Patel.
\newblock Deep image compositing.
\newblock In {\em WACV}, 2021.

\bibitem{deepblending}
Lingzhi Zhang, Tarmily Wen, and Jianbo Shi.
\newblock Deep image blending.
\newblock In {\em WACV}, 2019.

\bibitem{lpips}
Richard Zhang, Phillip Isola, Alexei~A Efros, Eli Shechtman, and Oliver Wang.
\newblock The unreasonable effectiveness of deep features as a perceptual
  metric.
\newblock In {\em CVPR}, 2018.

\end{thebibliography}


\begin{thebibliography}{10}\itemsep=-1pt

\bibitem{blendedlatentdiffusion}
Omri Avrahami, Ohad Fried, and Dani Lischinski.
\newblock Blended latent diffusion.
\newblock {\em ArXiv}, abs/2206.02779, 2022.

\bibitem{btmodel1}
Ralph~Allan Bradley and Milton~E. Terry.
\newblock Rank analysis of incomplete block designs: I. the method of paired
  comparisons.
\newblock {\em Biometrika}, 39:324, 1952.

\bibitem{cdtnet}
Wenyan Cong, Xinhao Tao, Li Niu, Jing Liang, Xuesong Gao, Qihao Sun, and Liqing
  Zhang.
\newblock High-resolution image harmonization via collaborative dual
  transformations.
\newblock In {\em CVPR}, 2022.

\bibitem{dovenet}
Wenyan Cong, Jianfu Zhang, Li Niu, Liu Liu, Zhixin Ling, Weiyuan Li, and Liqing
  Zhang.
\newblock {DoveNet}: Deep image harmonization via domain verification.
\newblock In {\em CVPR}, 2020.

\bibitem{qualityscore}
Shuyang Gu, Jianmin Bao, Dong Chen, and Fang Wen.
\newblock Giqa: Generated image quality assessment.
\newblock In {\em ECCV}, 2020.

\bibitem{pctnet}
Julian Jorge~Andrade Guerreiro, Mitsuru Nakazawa, and Bj\"orn Stenger.
\newblock Pct-net: Full resolution image harmonization using pixel-wise color
  transformations.
\newblock In {\em CVPR}, 2023.

\bibitem{fid}
Martin Heusel, Hubert Ramsauer, Thomas Unterthiner, Bernhard Nessler, and Sepp
  Hochreiter.
\newblock Gans trained by a two time-scale update rule converge to a local nash
  equilibrium.
\newblock In {\em NIPS}, 2017.

\bibitem{segmentanything}
Alexander Kirillov, Eric Mintun, Nikhila Ravi, Hanzi Mao, Chloe Rolland, Laura
  Gustafson, Tete Xiao, Spencer Whitehead, Alexander~C. Berg, Wan-Yen Lo, Piotr
  Doll{\'a}r, and Ross Girshick.
\newblock Segment anything.
\newblock {\em arXiv:2304.02643}, 2023.

\bibitem{btmodel2}
Wei-Sheng Lai, Jia-Bin Huang, Zhe Hu, Narendra Ahuja, and Ming-Hsuan Yang.
\newblock A comparative study for single image blind deblurring.
\newblock In {\em CVPR}, 2016.

\bibitem{stgan}
Chen{-}Hsuan Lin, Ersin Yumer, Oliver Wang, Eli Shechtman, and Simon Lucey.
\newblock {ST-GAN:} spatial transformer generative adversarial networks for
  image compositing.
\newblock In {\em CVPR}, 2018.

\bibitem{sdedit}
Chenlin Meng, Yutong He, Yang Song, Jiaming Song, Jiajun Wu, Jun-Yan Zhu, and
  Stefano Ermon.
\newblock Sdedit: Guided image synthesis and editing with stochastic
  differential equations.
\newblock In {\em ICLR}, 2021.

\bibitem{poissonblending}
Patrick P{\'e}rez, Michel Gangnet, and Andrew Blake.
\newblock Poisson image editing.
\newblock {\em SIGGRAPH}, 2003.

\bibitem{clip}
Alec Radford, Jong~Wook Kim, Chris Hallacy, Aditya Ramesh, Gabriel Goh,
  Sandhini Agarwal, Girish Sastry, Amanda Askell, Pamela Mishkin, Jack Clark,
  Gretchen Krueger, and Ilya Sutskever.
\newblock Learning transferable visual models from natural language
  supervision.
\newblock In {\em ICML}, 2021.

\bibitem{objectstitch}
Yi-Zhe Song, Zhifei Zhang, Zhe~L. Lin, Scott~D. Cohen, Brian~L. Price, Jianming
  Zhang, Soo~Ye Kim, and Daniel~G. Aliaga.
\newblock Objectstitch: Generative object compositing.
\newblock In {\em CVPR}, 2023.

\bibitem{dccf}
Ben Xue, Shenghui Ran, Quan Chen, Rongfei Jia, Binqiang Zhao, and Binqiang
  Zhao.
\newblock Dccf: Deep comprehensible color filter learning framework for
  high-resolution image harmonization.
\newblock In {\em ECCV}, 2022.

\bibitem{paintbyexample}
Binxin Yang, Shuyang Gu, Bo Zhang, Ting Zhang, Xuejin Chen, Xiaoyan Sun, Dong
  Chen, and Fang Wen.
\newblock Paint by example: Exemplar-based image editing with diffusion models.
\newblock In {\em CVPR}, 2023.

\bibitem{sfgan}
Fangneng Zhan, Hongyuan Zhu, and Shijian Lu.
\newblock Spatial fusion {GAN} for image synthesis.
\newblock In {\em CVPR}, 2019.

\bibitem{deepblending}
Lingzhi Zhang, Tarmily Wen, and Jianbo Shi.
\newblock Deep image blending.
\newblock In {\em WACV}, 2019.

\end{thebibliography}
}

\end{document}

\title{Supplementary for ControlCom: Controllable Image Composition using Diffusion Model}

\author{
Bo Zhang$^{1}$
\quad Yuxuan Duan$^{1}$
\quad Jun Lan$^{2}$
\quad Yan Hong$^{2}$
\\
\quad Huijia Zhu$^{2}$
\quad Weiqiang Wang$^{2}$
\quad Li Niu\thanks{Corresponding author} $^{1}$ \\
\quad $^1$ Shanghai Jiao Tong University \\
{\tt\small\{bo-zhang, sjtudyx2016, ustcnewly\}@sjtu.edu.cn} \and
\quad $^2$ Ant Group \\
{\tt\small\{yelan.lj, ruoning.hy, huijia.zhj, weiqiang.wwq\}@antgroup.com}
}

\maketitle

In this document, we provide additional materials to supplement our main text. We will first provide more details and explanations of training data preparation in Sec.~\ref{sec:data_prepare}.
In Sec.~\ref{sec:utility_controllability}, we will demonstrate the utility of controllable image composition. In Sec.~\ref{sec:ablation_study}, we will validate the effectiveness of various components in our model through both qualitative and quantitative results. Next, more visual results, including qualitative comparison of different baselines and controllable image composition, will be provided in Sec.~\ref{sec:more_visual_results}, which can highlight the advantages of our method over existing ones. In Sec.~\ref{sec:user_study}, we will conduct user studies on individual tasks for subjective evaluation and show some visualization results. Finally, in Sec.~\ref{sec:limitations}, we will analyze the limitation of the proposed method by several failure cases.

\section{Training Data Preparation}
\label{sec:data_prepare}
In Sec. 4.3 of the main text, we have briefly described the pipeline of the data augmentation and training sample generation. Here we provide more details about this process.
\subsection{Data Augmentation}
\noindent\textbf{Composite Image Augmentation.} We apply both random crop and illumination augmentation to composite image. When performing random crop, we ensure that the foreground object is always contained in the crop window. For illumination augmentation, we realize it by color jitter, which randomly changes the brightness, saturation, contrast and hue within a range of [0.8, 1.2], [0.8, 1.2], [0.8, 1.2], and [-0.05, 0.05], respectively.

\noindent\textbf{Foreground Image Augmentation.} The foreground image is cropped from the same source image as the composite image. To simulate real use-case scenarios, where input foreground is often from other source images, we apply a strategy called background swap to manipulate the non-foreground region of each foreground image. Specifically, we crop a patch of the same size as the given foreground image from another randomly selected source image and paste the foreground object at the center of the patch. We then perform illumination augmentation and geometry augmentation on the foreground successively, producing $\mathbf{I}^u_f$ and $\mathbf{I}^g_f$, respectively. The illumination augmentation is implemented by color jitter with the same parameters as the composite image. The geometry augmentation consists of horizontal flip with probability 0.2, random rotation within the range [$-$20, 20], and randomly perturbing the four corner points of the foreground object by perspective transformation. Note the illumination augmentation is performed on the whole image, while the geometry augmentation is only applied to the foreground object to avoid the geometric distortion of background in the foreground. After that, we use random blur with probability of 0.3 to further enhance model robustness.

\begin{figure*}[t]
    \begin{center}
    \includegraphics[width=1\linewidth]{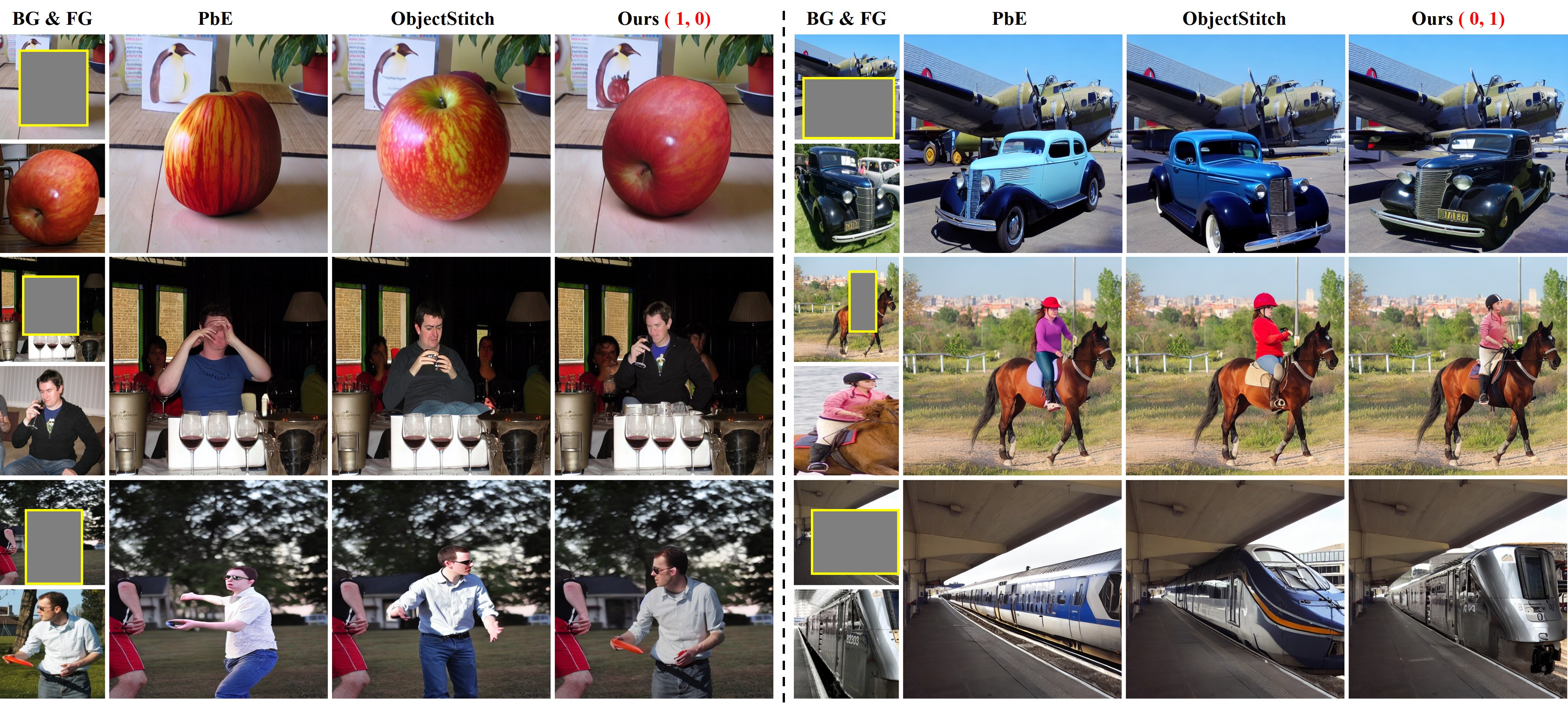}
    \end{center}
    \caption{Some examples highlighting the advantages of controllable image composition over existing generative composition methods, \ie, PbE~\cite{paintbyexample} and ObjectStitch~\cite{objectstitch}.}
    \label{fig:controllability_necessity}
\end{figure*}

\subsection{Training Sample Generation}
In Sec. 4.3 of the main text, we give the strategy of collecting various tuples for the four tasks. Here we present more elaborate discussion on the motivation behind the strategy.       

\noindent\textbf{Image Blending with Indicator (0,0).} We only apply illumination augmentation to the foreground image, \ie, $\mathbf{I}_f = \mathbf{I}^u_f$, and replace the foreground object in $\mathbf{I}^u_c$ with that in $\mathbf{I}^u_f$ to get pseudo ground-truth composite $\mathbf{I}^n_c$. In the pseudo ground-truth composites, foreground object usually has inconsistent illumination with background scene. If the pseudo ground-truth composites always have consistent illumination, then the model tends to alter the foreground illumination to fit the background during inference time. In contrast, by using inconsistent pseudo ground-truth, we enforce model to maintain the foreground illumination during image synthesis. Moreover, by training on image blending task, the fidelity of our model can benefit from learning to reconstruct the foreground object on background scene accurately.

\noindent\textbf{Image Harmonization with Indicator (1,0).} We set $\mathbf{I}_f = \mathbf{I}^u_f, \mathbf{I}_c = \mathbf{I}^u_c$. The pseudo ground-truth composite image blends the foreground and background seamlessly. In this case, we feed into inharmonious background and foreground, and guide the model to adjust foreground illumination, generating a harmonious image close to the pseudo ground-truth.

\noindent\textbf{View Synthesis with Indicator (0,1).} In this case, the controllable generator is expected to synthesize foreground object with novel viewpoint compatible with the given background, which typically involves geometric transformation. To simulate this, we set $\mathbf{I}_f = \mathbf{I}^g_f, \mathbf{I}_c=\mathbf{I}^n_c$. Compared with the foreground $\mathbf{I}_f$, the foreground of $\mathbf{I}_c$ has consistent illumination yet discrepant geometric conditions (\eg, pose and scale). In this way, the model is expected to change the foreground pose while remaining its illumination.  

\noindent{\textbf{Generative Composition with Indicator (1,1).}} we set $\mathbf{I}_f = \mathbf{I}^g_f, \mathbf{I}_c = \mathbf{I}^u_c$. Therefore, the pseudo ground-truth composite images have consistent illumination and geometry, which are generally different from the input foreground. By feeding inconsistent background and foreground, we encourage the model to change both illumination and pose of foreground to generate realistic composite image.

Moreover, we provide details about computing the evaluation metrics introduced in the Sec. 5.2 of the main text. To measure the similarity between the given and synthesized foreground object, we crop out foreground patch from generated composite and mask the non-object region to compute CLIP score~\cite{clip} ($\mathrm{CLIP_{fg}}$) with the masked input foreground, in which we estimate the mask of foreground object using SAM~\cite{segmentanything}. For evaluating generative quality, we adopt Fréchet Inception Distance (FID)~\cite{fid} and Quality Score (QS)~\cite{qualityscore} to evaluate the authenticity of both generated composite images and foreground regions following~\cite{paintbyexample}. Specifically, we compute FID and QS between synthesized images and all images from COCO test set. The $\mathrm{FID_{fg}}$ is calculated between synthesized/real foreground patches, where real foreground patches are collected from all source images of COCOEE dataset~\cite{paintbyexample}.

\begin{figure*}[t]
    \begin{center}
    \includegraphics[width=1\linewidth]{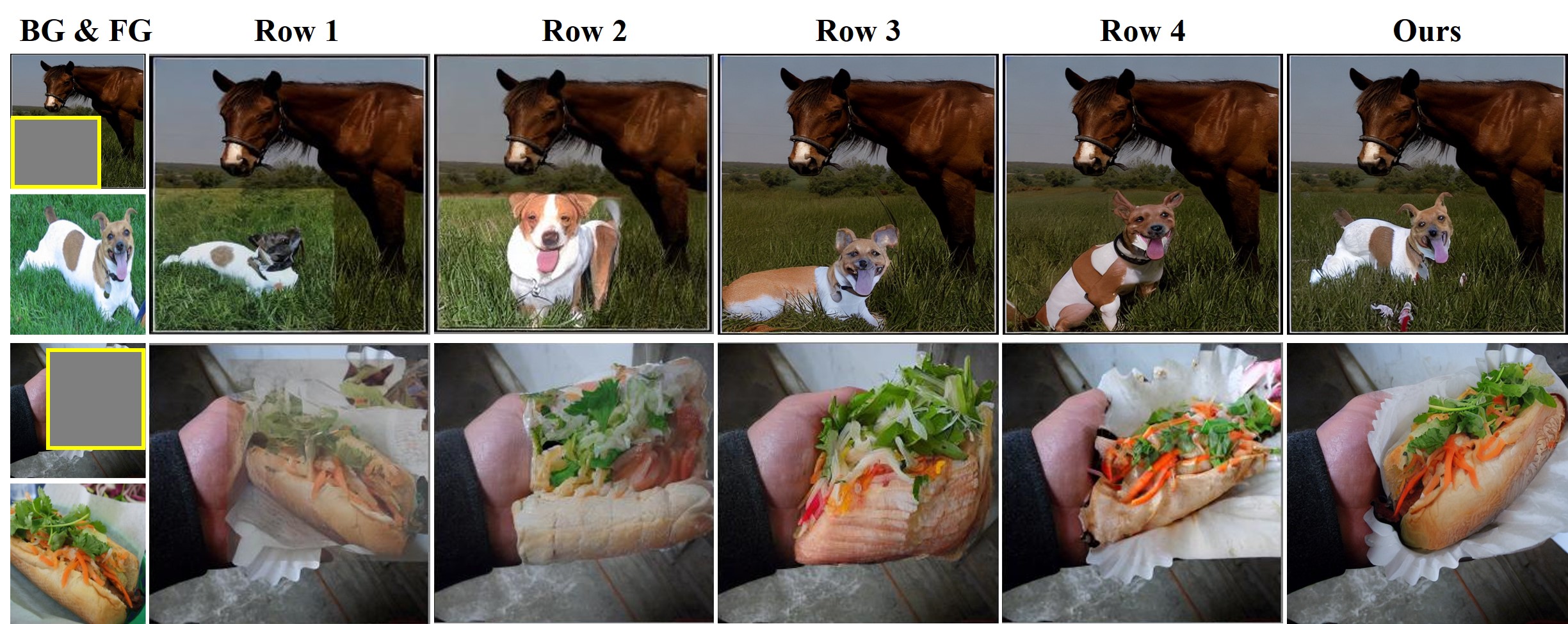}
    \end{center}
    \caption{Visual results of the ablation study in Table~\ref{tab:ablation_study}.}
    \label{fig:ablation_study}
\end{figure*}

\section{The Utility of Controllable Image Composition}
\label{sec:utility_controllability}
In Sec. 1 of the main text, we discuss the necessity of developing controllable image composition, \ie, enabling user to preserve some attributes (\eg, illumination and pose) of foreground object when generating composite images. As an example, in the second row of Figure 6 in the main text, the dog in the foreground image is compatible with the background scene, thus user probably prefers to preserve the pose in the composite image. In this Section, we provide additional examples to demonstrate the utility of controllable image composition.

On the left of Figure~\ref{fig:controllability_necessity}, we present some examples where the pose of foreground is already congruous with the background scene. In these images, users may expect to maintain the foreground pose while harmonizing its appearance. However, existing generative composition methods, \ie, PbE~\cite{paintbyexample} and ObjectStitch~\cite{objectstitch}, usually change the foreground pose in an uncontrollable manner. By taking the last row as an example, user may want to retain the pose of the person holding the Frisbee in the composite image, which have been altered in the results of PbE and ObjectStitch. In contrast, when setting indicator as (1,0), our model enables user to preserve the foreground pose while performing illumination adjustment, yielding faithful composite images in the last column.       

On the right of Figure~\ref{fig:controllability_necessity}, we show several examples for another situation, in which the foreground image has illumination condition roughly matching the background image. Therefore, user may hope to preserve the inherent illumination of foreground to avoid unnecessary and even unreasonable color change, while only adjusting its pose. When using PbE~\cite{paintbyexample} and ObjectStitch~\cite{objectstitch}, they may severely alter the foreground color in an undesirable manner. For instance, in the top-right example of Figure~\ref{fig:controllability_necessity}, the color of the cars generated by the baselines obviously deviates from that of the input foreground. Differently, our method can handle such cases well by setting the indicator to (0,1). In particular, our method can generate realistic composite images without unnecessary and even unreasonable color change. 

\begin{table}
\centering
% \resizebox{\linewidth}{!}{
\begin{tabular}{l|ccc|ccc}
\hline
& GF & DA & LE  &$\mathrm{CLIP_{fg}}$↑ & FID↓ & QS↑ \\
\hline \hline
1 &AT & & & 85.52 & 3.74 & 71.29 \\
2 &+ & & & 83.55  & 3.58 & 73.07\\
3 &+ &+ & & 84.39 & 3.41 & 76.88 \\
4 &+ &+ &w/o FM & 86.76	& 3.33	& 77.14 \\
5 &+ &+ &+ & \textbf{88.31} & \textbf{3.19}	& \textbf{77.84} \\
\hline
\end{tabular}
% }
\caption{Ablation study of the proposed method. GF: global fusion. LE: local enhancement. DA: training data augmentation. AT: using all foreground tokens for global fusion. FM: feature modulation in Eqn.~(3) of the main text.}
\label{tab:ablation_study}
\end{table}

\begin{figure*}[p]
    \begin{center}
    \includegraphics[width=0.98\linewidth]{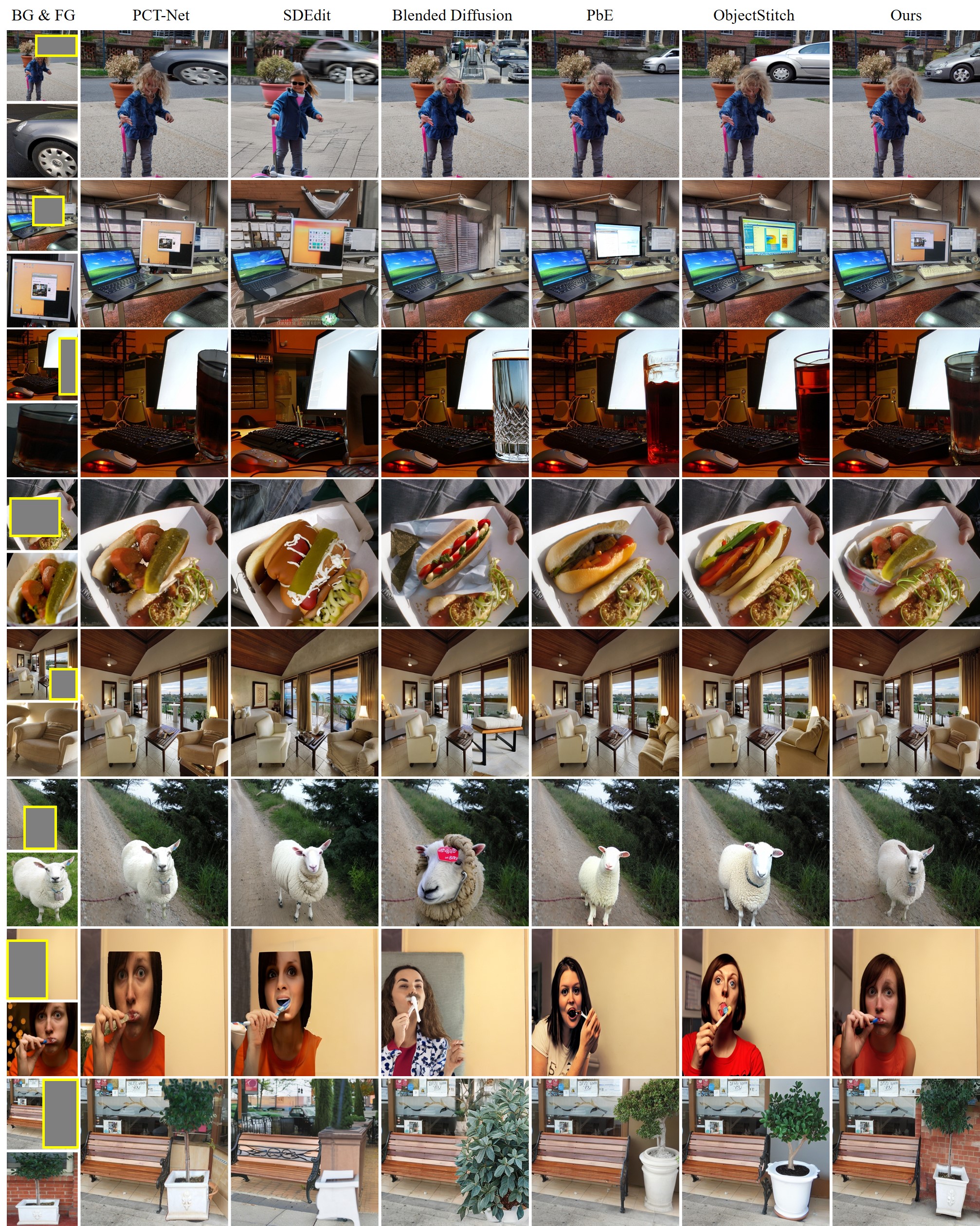}
    \end{center}
    \caption{Qualitative comparison on COCOEE dataset.}
    \label{fig:more_cocoee}
\end{figure*}

\begin{figure*}[p]
    \begin{center}
    \includegraphics[width=0.98\linewidth]{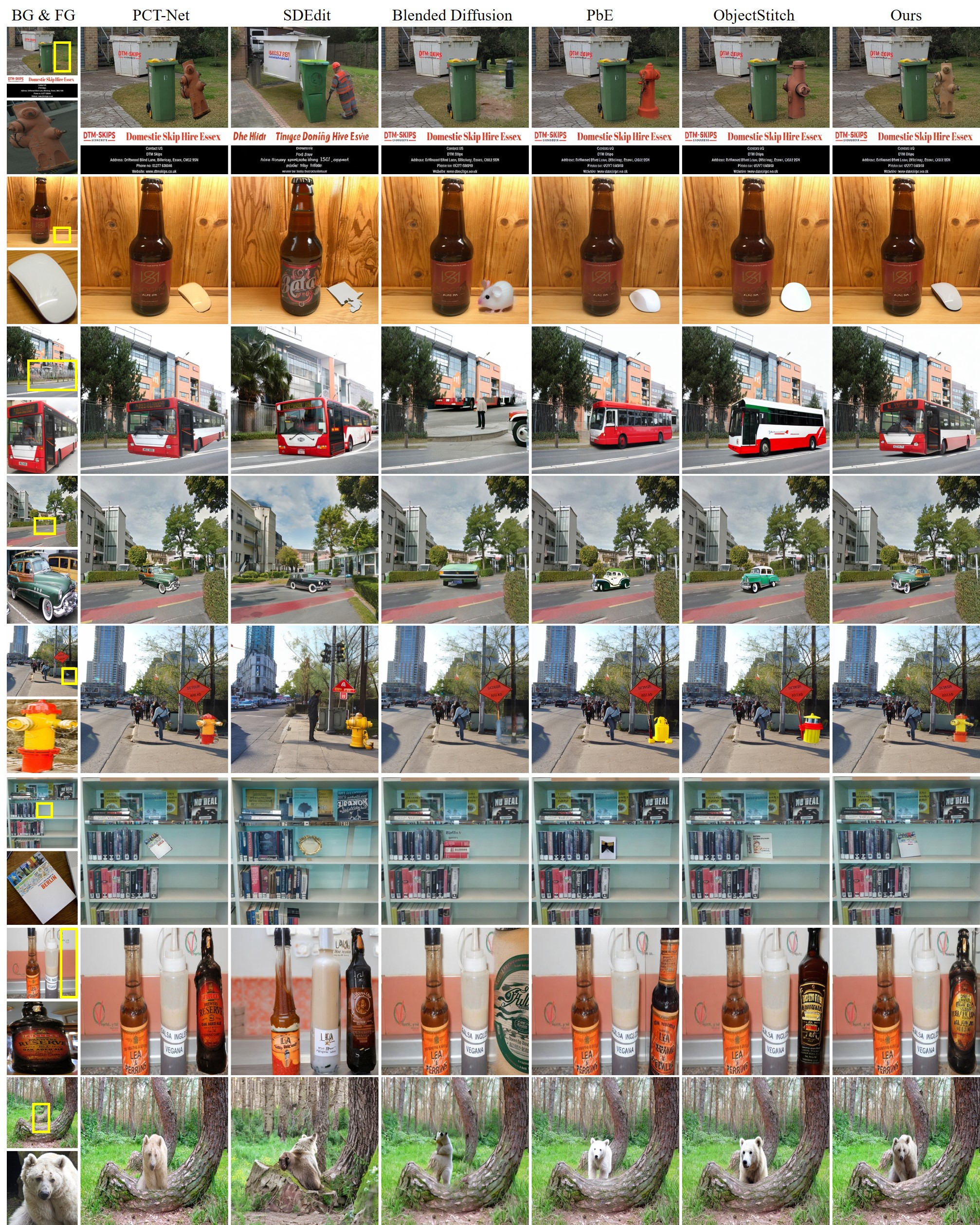}
    \end{center}
    \caption{Qualitative comparison on our FOSCom dataset.}
    \label{fig:more_foscom}
\end{figure*}

\begin{figure*}[p]
    \begin{center}
    \includegraphics[width=0.9\linewidth]{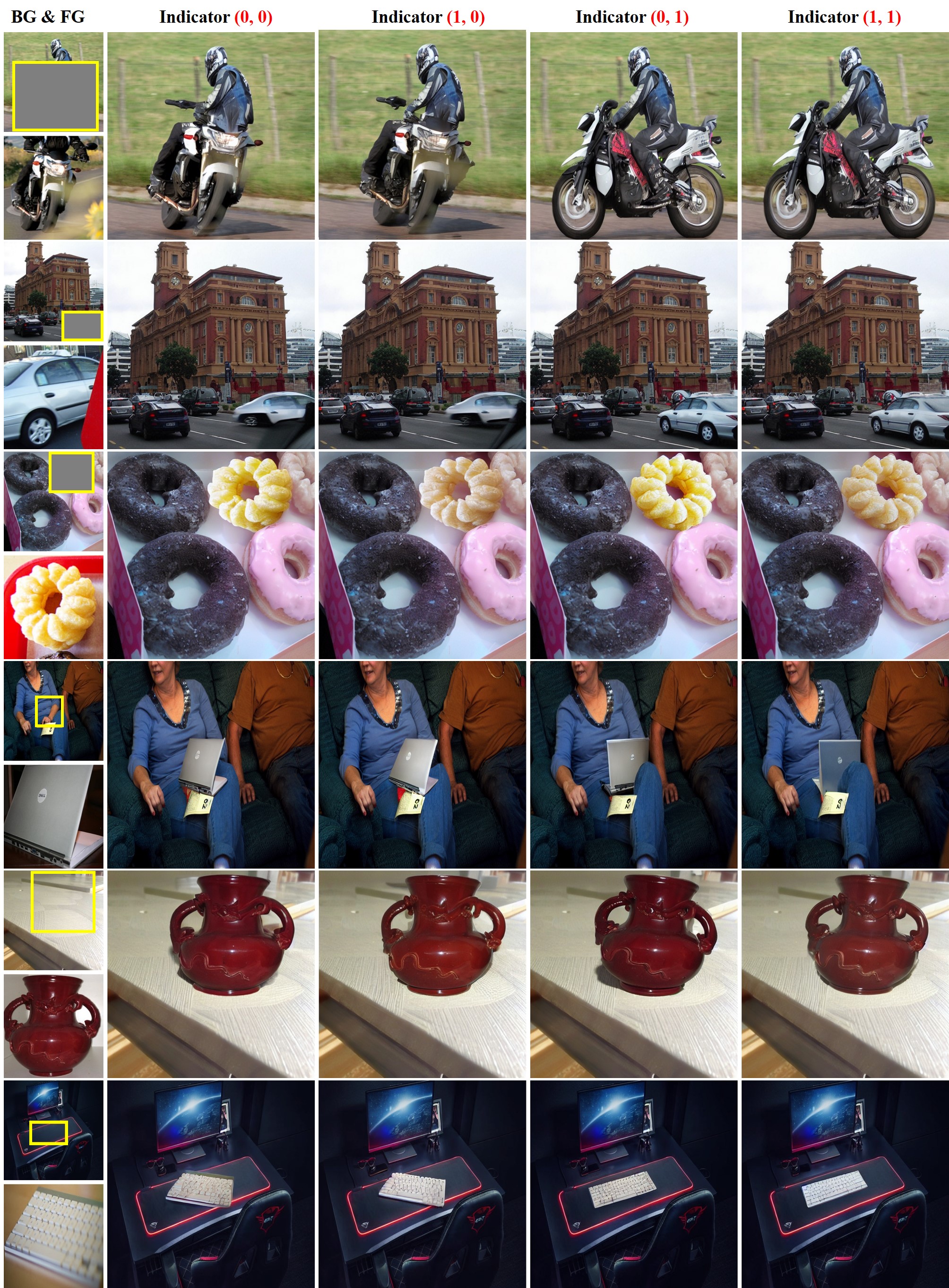}
    \end{center}
    \caption{Results of controllable image composition on COCOEE dataset (top half) and our FOSCom dataset (bottom half).}
    \label{fig:more_controllable}
\end{figure*}

\begin{table*}
\centering
% \resizebox{\linewidth}{!}{
\begin{tabular}{lc|lc|lcc}
\hline
\multicolumn{2}{c|}{Image Blending} &\multicolumn{2}{c|}{Image Harmonization} &\multicolumn{3}{c}{Generative Composition}\\
Method &B-T Score↑ &Method &B-T Score↑ &Method &Quality↓ &Fidelity↓ \\ 
\hline \hline
Poisson Blending &0.014 &-- &-- &-- &-- &--\\
Deep Blending &0.586 &DCCF &0.116 &-- &-- &--  \\
SDEdit &-0.570 &CDTNet &-0.297 &ObjectStitch &2.55 &2.26  \\
Blended Diffusion &-1.232 &PCT-Net &\textbf{0.270} &PbE &\textbf{1.72} &2.60 \\
\hline
Ours (Blend) &\textbf{1.201}  &Ours (Harm) &-0.089 &Ours (Comp) &1.73 &\textbf{1.14} \\
\hline
\end{tabular}
% } 
\caption{User study using our FOSCom dataset on individual tasks. For image blending and harmonization, we compute B-T score on the outputs of 100 samples. For generative composition, we compute average ranking score of image quality and foreground fidelity for 640 samples. Boldface indicates the best results.}
\label{tab:user_study}
\end{table*}

\section{Ablation Study}
\label{sec:ablation_study}
We conduct ablation studies on COCOEE dataset~\cite{paintbyexample} to evaluate the effectiveness of various components in our method, including Global Fusion (GF) module, Local Enhancement (LE) module, and training data augmentation. We show the numerical results in Table~\ref{tab:ablation_study}, together with the visual results in Figure~\ref{fig:ablation_study}. 
We start from a simple solution, which replaces the text embedding of text-guided diffusion model with all tokens of foreground as condition information, and gradually add proposed modifications to build our model. Note that we generate results for all rows in Table~\ref{tab:ablation_study} with indicator (1,1) except for row 1 and row 2. 
In row 1 and 2, we train the model without using indicator or any data augmentation. During training, the source image is directly adopted as pseudo ground-truth composite, \ie, $\mathbf{I}_c=\mathbf{I}_s$ and the foreground region cropped from $\mathbf{I}_s$ is taken as $\mathbf{I}_f$. In such case, model cannot guarantee the control over foreground attributes. Furthermore, the model tends to recover $\mathbf{I}_c$ by simply copying and pasting $\mathbf{I}_f$ on the background image, resulting in the obvious copy-and-paste artifacts, as demonstrated in the second and third columns of Figure~\ref{fig:ablation_study}. More precisely, in row 1, we use both class token and patch tokens extracted by CLIP encoder~\cite{clip} to expand the length of global embedding from 1 to 257. This model generates extremely unnatural composite images with copy-and-paste artifacts, corresponding to high FID and low quality score. In row 2, we discard patch tokens and only use class token to generate global embedding. This modification slightly improves the generative quality (FID and QS) but reduces the similarity with input foreground.

To boost model generalization, we introduce the data augmentations together with four indicators in Sec.~4.3 of the main text to produce training samples, increasing image quality in row 3. The similar observation can be obtained from the comparison between the third and fourth columns in Figure~\ref{fig:ablation_study}.  
To enhance the foreground details, we further add two-stage fusion strategy with local enhancement module to row 3. Specifically, we first fuse local embedding $\mathbf{E}_l$ of foreground through plain cross attention (without using feature modulation) in row 4, greatly increasing the similarity between synthesized object and input foreground image. In this case, we directly use the synthesized foreground feature map $\tilde{\mathbf{F}}^l_i$ to update the global background feature $\mathbf{F}_i$ in Figure~3 of the main text. Then the feature modulation is adopted to build our full method in row 5, which further promotes the overall quality and fidelity of composite images and achieves the best performance. In the last three columns of Figure~\ref{fig:ablation_study}, it can be seen that the synthesized foreground is increasingly similar in appearance detail to the provided foreground, which also confirms the conclusion from Table~\ref{tab:ablation_study}.    

\section{More Visualization Results}
\label{sec:more_visual_results}
To better demonstrate the effectiveness of our method, we provide additional qualitative results of our method and baseline methods on the public COCOEE dataset~\cite{paintbyexample} and our FOSCom dataset in Figure~\ref{fig:more_cocoee} and Figure~\ref{fig:more_foscom}, respectively. The baseline methods include PCT-Net~\cite{pctnet}, SDEdit~\cite{sdedit}, Blended Diffusion~\cite{blendedlatentdiffusion}, PbE~\cite{paintbyexample}, and ObjectStitch~\cite{objectstitch}. Given a pair of background and foreground images, we show the composite images that are generated by different methods, in which our results are produced by setting indicator as (1,1). These visualized results demonstrate that our method can generally synthesize plausible composite images with more faithful foreground than other approaches. 

We also present the visual results of our four versions in Figure~\ref{fig:more_controllable}, in which four composite images are generated from the same initial Gaussian noise using different indicators. When setting indicator to (0,0), our Blending version attempts to reconstruct the foreground object and blends it into the background image. By feeding indicator (1,0) and (0,1), our Harmonization version and ViewSynthesis version support to adaptively adjust foreground illumination and pose to fit the given background scene, respectively. Finally, our Composition version with indicator (1,1) can create high-quality composite images, by virtue of performing illumination adjustment and novel view synthesis simultaneously. These examples further demonstrate the strong controllability of the proposed method on foreground illumination and pose.    

 \begin{figure*}[t]
    \begin{center}
    \includegraphics[width=0.85\linewidth]{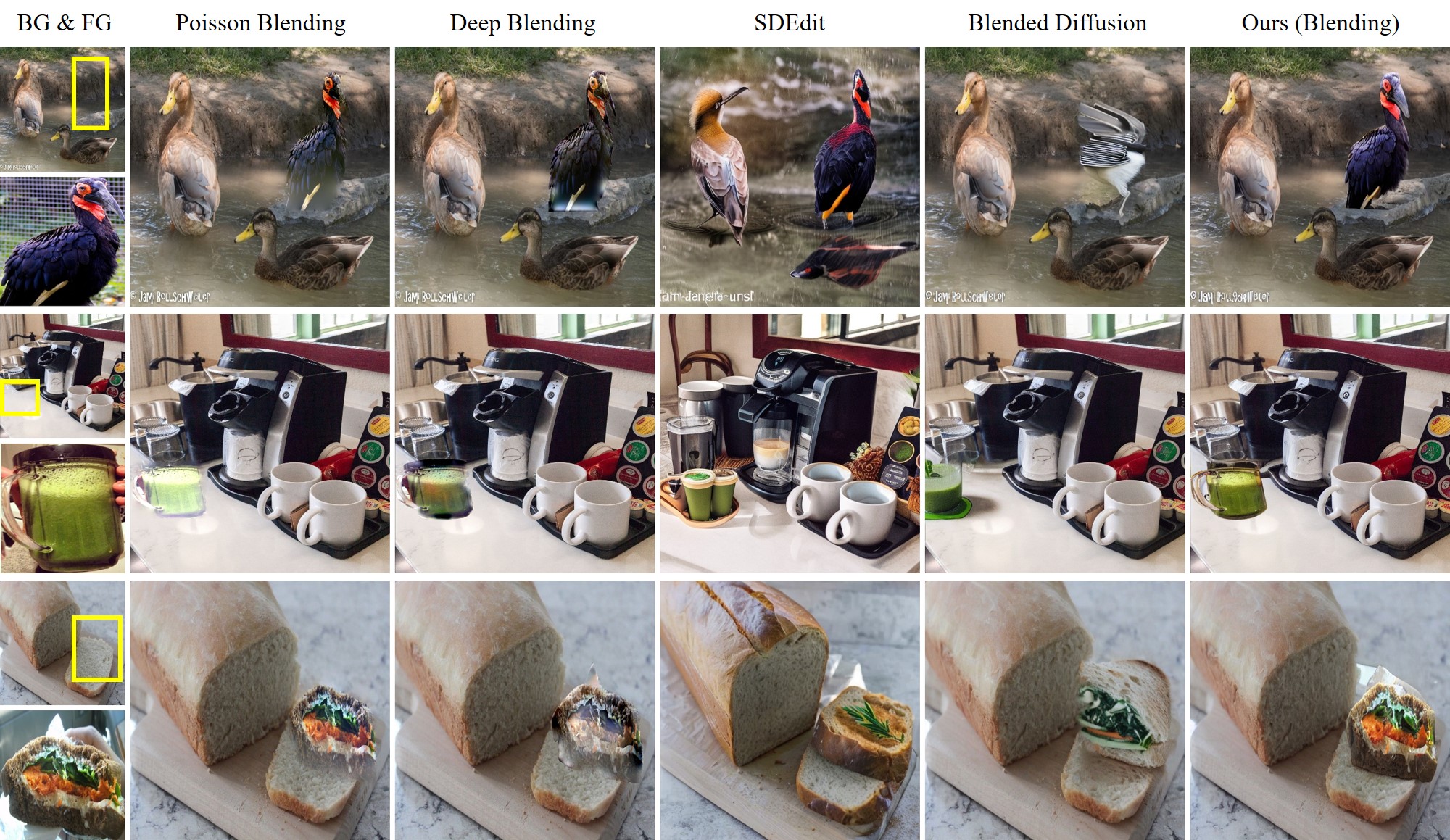}
    \end{center}
    \caption{Qualitative comparison on image blending. The baseline methods include Poisson blending~\cite{poissonblending}, deep image blending~\cite{deepblending}, SDEdit~\cite{sdedit}, and blended latent diffusion~\cite{blendedlatentdiffusion}.}
    \label{fig:blending}
\end{figure*}

\begin{figure*}[t]
    \begin{center}
    \includegraphics[width=0.85\linewidth]{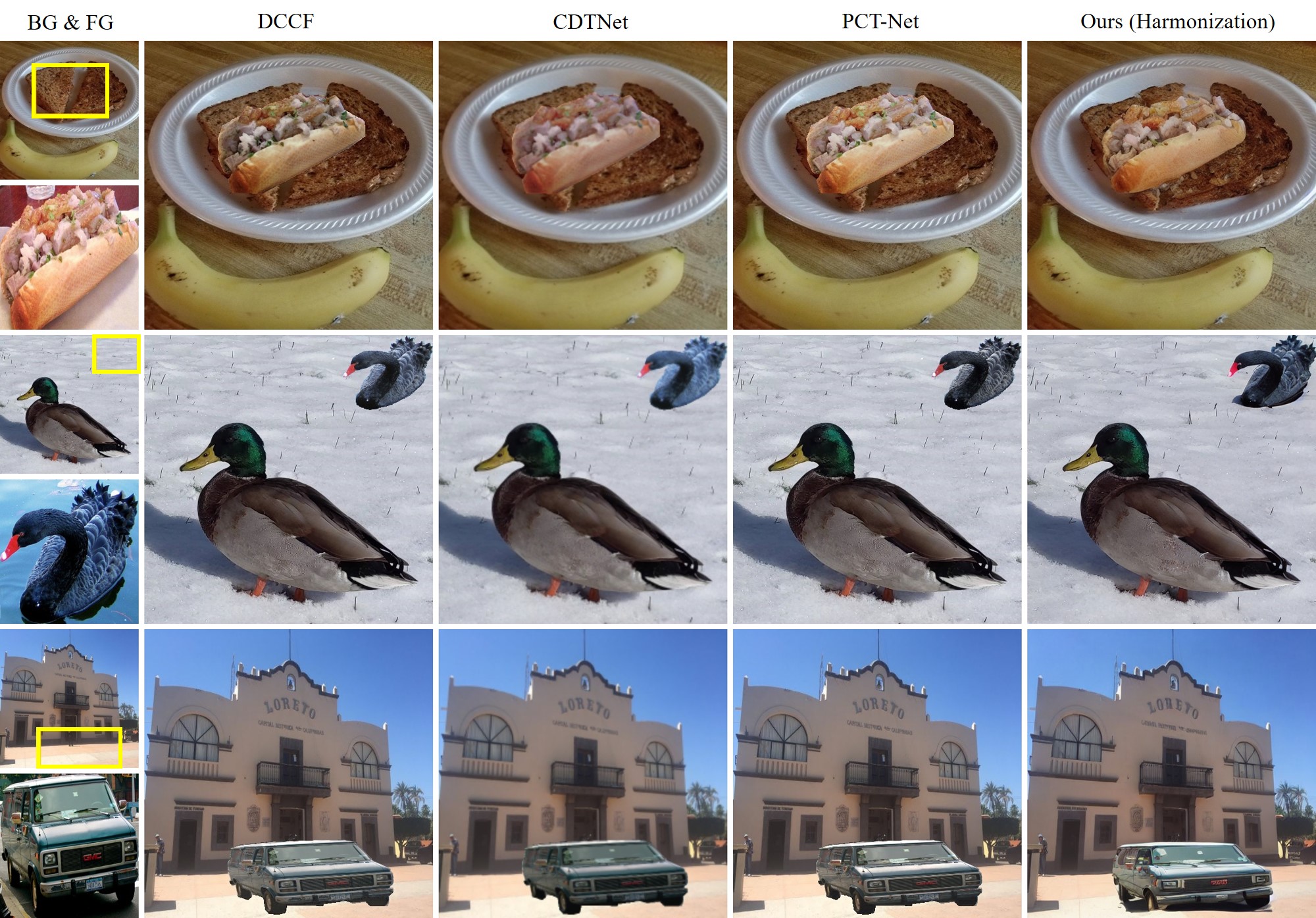}
    \end{center}
    \caption{Qualitative comparison on image harmonization. The baseline methods include: DCCF~\cite{dccf}, PCT-Net~\cite{pctnet}, and CDTNet~\cite{cdtnet}.}
    \label{fig:harmonization}
\end{figure*}

\section{User Study on Individual Tasks}
\label{sec:user_study}
We conduct user study to compare our method with other approaches on individual tasks, including image blending, image harmonization, and generative composition. For view synthesis, existing spatial transformation methods~\cite{stgan,sfgan} relying on perspective transformation struggle to tackle complex viewpoint transformation, leading to the shortage of competitive baselines. 

On COCOEE and FOSCom datasets, we do not have ground-truth for image blending and harmonization tasks, and the evaluation metrics like $\mathrm{CLIP}$ and $\mathrm{FID}$ are not suitable for these tasks. Therefore, we conduct user study on each of the three tasks for subjective evaluation and report results in Table~\ref{tab:user_study}.
For image blending and image harmonization, we carry out user study following~\cite{dovenet}, in which we employ 100 evaluation tuples from our FOSCom dataset. For each tuple, we provide a background image, a foreground image, and their composite image produced by Inpaint\&Paste (see Sec. 5.3 of the main text). 

\begin{figure}[t]
    \begin{center}
    \includegraphics[width=1\linewidth]{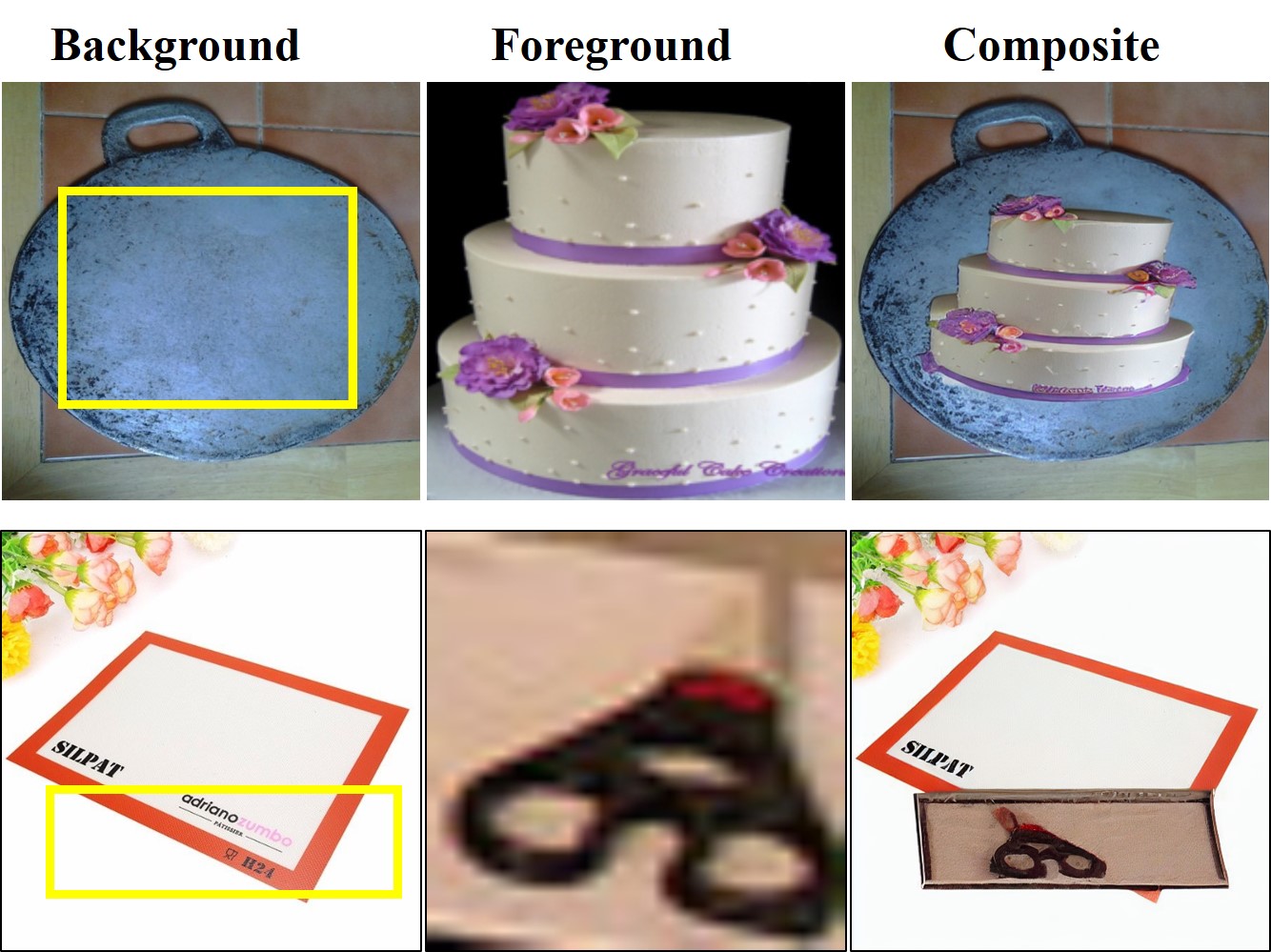}
    \end{center}
    \caption{Some failure cases of our method.}
    \label{fig:failure_cases}
\end{figure}

For image blending, we choose Poisson blending~\cite{poissonblending}, deep image blending~\cite{deepblending}, SDEdit~\cite{sdedit}, and blended latent diffusion~\cite{blendedlatentdiffusion} as baselines. Given a test tuple, we can obtain 5 outputs including the results of our Blending version. Based on 5 outputs, we construct pairs of outputs and invite 50 human raters to choose the one with more natural boundary between foreground and background. A total of 50000 pairwise results are collected for all 100 pairs of background/foreground. Finally, we use the Bradley-Terry model (B-T model)~\cite{btmodel1,btmodel2} to calculate the global ranking score for each method. From the left part of Table~\ref{tab:user_study}, we observe that conventional methods (Poisson blending~\cite{poissonblending} and deep image blending~\cite{deepblending}) are generally better than text-guided image generation/editing methods (deep image blending~\cite{deepblending}, SDEdit~\cite{sdedit}), on account of the limited representation of text information. Among these methods, our Blending version achieves the highest B-T score. We also provide several examples to visualize the comparison results in Figure~\ref{fig:blending}, which can verify the results in Table~\ref{tab:user_study}.  

 For image harmonization, we compare our Harmonization with DCCF~\cite{dccf}, PCT-Net~\cite{pctnet}, and CDTNet~\cite{cdtnet}. Similar to image blending, we select paired results from 4 outputs including the results of our Harmonization version, leading to total of 30000 pairwise results, and the raters are requested to choose the more harmonious one. From the middle part of Table~\ref{tab:user_study}, it can be seen that our method is surpassed by the state-of-the-art approaches~\cite{dccf,pctnet} specialized for this task, which typically learn from specific datasets. Despite that, our method is preferred by more participants than CDTNet~\cite{cdtnet} and is able to output harmonious composite images in common cases (see Figure~\ref{fig:harmonization}). Furthermore, as illustrated in the last two rows of Figure~\ref{fig:harmonization}, the proposed model does not only adjust the illumination of foreground, but also synthesizes plausible shadow for the foreground, which is missing in the results of other conventional methods.
 
 For generative composition, instead of using B-T score, we adopt average ranking score to measure image quality and foreground fidelity following~\cite{paintbyexample}. Specifically, we use all 640 pairs of background and foreground images from FOSCom dataset in the study. For each pair, we generate three composite results using three generative composition approaches, \ie, PbE~\cite{paintbyexample}, ObjectStitch~\cite{objectstitch}, and our Composition version. Thus we obtain 640 groups of image with each group containing two inputs and three outputs. All these results in each group are present side-by-side and in a random order to 50 human participants. Participants rank the score from 1 to 5 (1 is the best, 5 is the worst) on the overall quality and fidelity of composite images independently, without time limitation. Finally, we report the average ranking score on the right of Table~\ref{tab:user_study}, from which we have similar observation as in Table 1 of the main text. In particular, given the comparable image quality of our method, raters prefer our results more than others considering the high fidelity of ours.  

\section{Limitations}
\label{sec:limitations}
The proposed method is generally able to produce realistic composite images with high fidelity, but it still suffers from several limitations that probably lead to implausible images. In the top row of Figure~\ref{fig:failure_cases}, given a side view of the cake, the model ought to synthesize the top view of the cake that properly fits the background scene, but it fails to do so. This reveals that it is challenging to synthesize novel view for the provided foreground, when the novel view and the current view have little or no overlapping. A possible solution is training on image pairs containing the same object captured from diverse camera views, so that the model might get better at synthesizing a novel view with huge view discrepancy. 

Another noteworthy limitation is caused by low-quality input images, \eg, blurred and dim foreground. During training, we have already performed various data augmentation on input images, simulating the practical situations of low-quality inputs. However, our model may still output unnatural composite images with artifacts given the low-quality foreground. We provide one example in the bottom row of Figure~\ref{fig:failure_cases}. To address this issue, we may need to collect or synthesize more foreground images in such extreme cases, which could be used to learn a more robust generator.

%%%%%%%%% REFERENCES
{\small
\bibliographystyle{ieee_fullname}
\bibliography{egbib}
}